\documentclass[runningheads]{llncs}
\usepackage[T1]{fontenc}
\usepackage{hyperref}
\usepackage{graphicx}
\usepackage{bm}
\usepackage{bbm}
\usepackage{amsmath}
\usepackage{amssymb}
\usepackage{booktabs}
\usepackage{multirow}
\usepackage{caption}
\usepackage{subcaption}
\usepackage{xcolor}
\usepackage{marvosym}

\usepackage{todonotes}

\begin{document}
\title{General Vision Encoder Features as Guidance in Medical Image Registration}

%
%
\author{Fryderyk Kögl \inst{1,2,3}*\textsuperscript{(\Letter)}  \and 
Anna Reithmeir\inst{2,3,4}*\textsuperscript{(\Letter)} \and 
Vasiliki Sideri-Lampretsa\inst{2} \and 
Ines Machado\inst{5,6} \and 
Rickmer Braren \inst{1,7} \and 
Daniel Rückert \inst{1,2,9} \and 
Julia A. Schnabel \inst{2,3,8} \and  
Veronika A. Zimmer\inst{1,2}} 

\authorrunning{F. Kögl, A. Reithmeir et al.}

\institute{School of Medicine \& Health, Klinikum Rechts der Isar, Technical University of Munich, Germany \and 
School of Computation, Information \& Technology, Technical University of Munich, Germany\\ \email{\{fryderyk.koegl, anna.reithmeir\}@tum.de} \and
Institute of Machine Learning in Biomedical Imaging, Helmholtz Munich, Germany \and 
Munich Center for Machine Learning, Germany \and
Cancer Research UK Cambridge Institute, University of Cambridge, UK \and
Department of Oncology, University of Cambridge, UK \and
German Cancer Consortium (DKTK), Partner Site Munich, Germany \and
School of Biomedical Engineering \& Imaging Sciences, King's College London, UK \and 
Department of Computing, Imperial College London, UK 
}
%
\maketitle              
\begin{abstract}
General vision encoders like DINOv2 and SAM have recently transformed computer vision. Even though they are trained on natural images, such encoder models have excelled in medical imaging, e.g., in classification, segmentation, and registration. However, no in-depth comparison of different state-of-the-art general vision encoders for medical registration is available.
In this work, we investigate how well general vision encoder features can be used in the dissimilarity metrics for medical image registration. We explore two encoders that were trained on natural images as well as one that was fine-tuned on medical data. 
We apply the features within the well-established B-spline FFD registration framework. In extensive experiments on cardiac cine MRI data, we find that using features as additional guidance for conventional metrics improves the registration quality. The code is available at \href{https://github.com/compai-lab/2024-miccai-koegl}{github.com/compai-lab/2024-miccai-koegl}.
\end{abstract}

\section{Introduction}\label{sec:intro}
\let\thefootnote\relax\footnotetext{*Equal contribution}

Image registration is a crucial step in medical image analysis that enables the alignment of corresponding anatomical or functional regions across multiple images, regardless of modality or differences between time points. 
This process creates a unified and accurate representation of patient data, facilitating enhanced diagnostic accuracy, improved treatment planning, and disease monitoring~\cite{Sotiras2013DeformableMI,haskins2020}. 

\begin{figure}[t]
    \centering
    \includegraphics[width=0.9\textwidth]{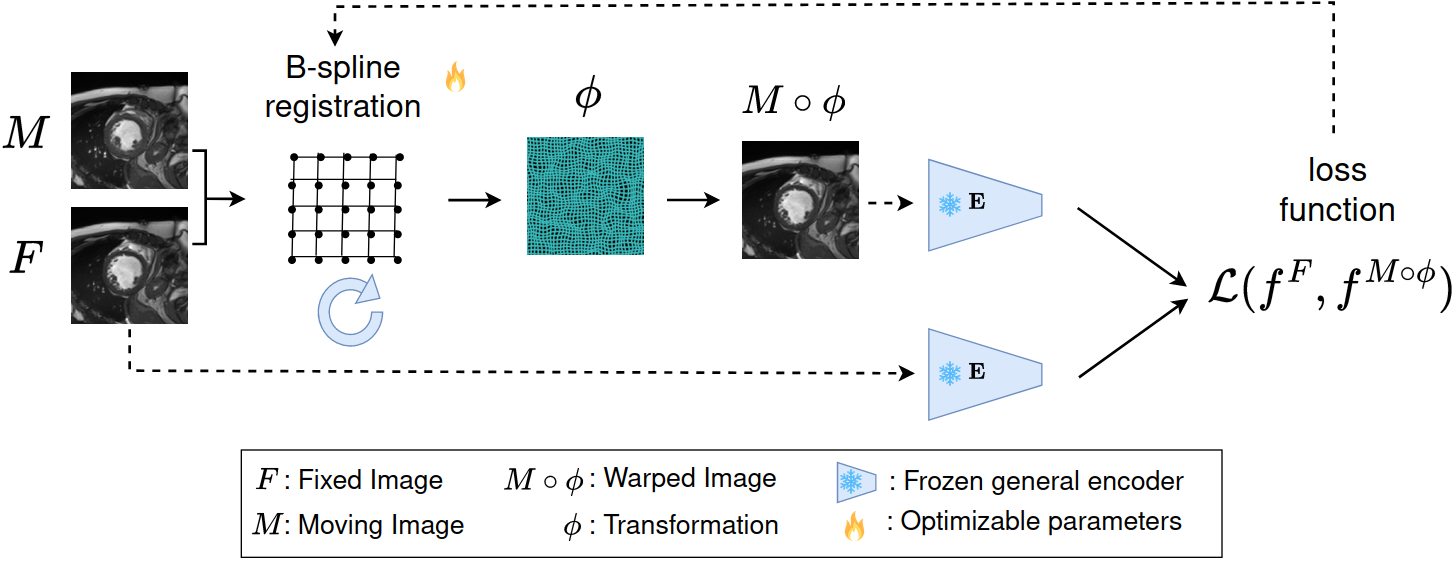}
    \caption{Overall framework: We explore how different pre-trained natural image encoders can be used within the registration objective function. The fixed image and the warped moving image are encoded with the frozen encoders, and the distance between the features is measured. This distance measure term is then used in the objective function of the iterative B-splines registration.}
    \label{fig:concept}
\end{figure}

Effective registration requires the assessment of the dissimilarity between images.
This is not trivial due to the highly complex, non-linear relationships arising from anatomical or functional differences.
As a result, choosing a distance measure is crucial and strongly influences the result.
Consequently, several measures have been tested in the context of registration. 
Pixel-based distance measures such as the Euclidian distance (mean squared error - MSE) between pixel intensities and the local or non-local normalized cross-correlation (NCC) are robust and common choices used in conventional~\cite{avants2008symmetric,avants2011reproducible,beg2005computing,rueckert1999nonrigid,thirion1998image,vercauteren2007non} and learning-based registration~\cite{balakrishnan2019voxelmorph,dalca2018unsupervised,dalca2019learning,chen2022transmorph,qiu2021learning}.
However, because registration is a non-convex optimization problem, these measures might be susceptible to local minima.
To account for that, hand-crafted local descriptor and histogram-based losses, e.g., modality independent neighborhood descriptor (MIND)~\cite{Heinrich2012MINDMI}, normalized gradient fields (NGFs)~\cite{Haber2006IntensityGB}, and mutual information (MI) and its variants~\cite{Studholme1999AnOI,wells1996multi,loeckx2009nonrigid}, have also been widely adopted, especially in the case of images with different contrasts. 
However, Mutual Information and its variants operate on intensity histograms; therefore, they are agnostic to the underlying geometry of the image structures.
Moreover, handcrafted features, derived from intensity and gradient information, may be sensitive to local details but lack global context, rendering them susceptible to local minima.

To account for the shortcomings of voxel-intensity measures~\cite{zhang2018unreasonable}, several works have studied the effects of data-driven measures in the context of registration \cite{haskins2019learning,Simonovsky2016ADM,Czolbe2020DeepSimSS,pielawski2020comir,sideri2023mad}.
However, each \textit{learned} measure comes with certain shortcomings. 
Haskins et al.~\cite{haskins2019learning} require ground truth transformations. 
Czolbe et al.~\cite{Czolbe2020DeepSimSS} propose to pre-train a feature extractor on a supplementary segmentation task and then use the semantic features to drive the optimization of a learning-based registration model for mono-modal registration. 
In contrast, \cite{pielawski2020comir,sideri2023mad} are tested only in the context of rigid/affine registration.
However, all the above methods need retraining to apply them to a new modality or image domain, rendering them impractical for most clinical scenarios.


Building upon the success of feature-based measures, we explore the potential of using features learned by large vision encoders to drive deformable image registration.
General vision encoders are deep learning models designed to learn semantic information from visual input such as images~\cite{bordes2024introduction,zhou2024towards,jin2024efficient,wadekar2024evolution} and videos~\cite{madan2024foundation}, and they can be applied to various downstream tasks once they have been trained. 
These general vision encoders provide versatile, rich feature embeddings, leveraging large, unlabeled datasets to enable task-agnostic performance, eliminating the need for task-specific data.
Additionally, they allow for seamless adaptation to new tasks without retraining, which is very important in the medical image domain, where data can be scarce.
For this reason, general vision encoders are being investigated for a wide range of applications~\cite{azad2023,cui2024,denner2024,huix2024,rabbani2024,zhang2024,blankemeier2024merlin}.
Nonetheless, these encoders have not been extensively explored in the context of medical image registration, with DINO-Reg~\cite{song2024general} to be the only work that explores the registration of DINO-v2~\cite{oquab2023dinov2} features.
In this work, instead of estimating the deformation field between feature maps as in~\cite{oquab2023dinov2}, we propose to utilize these semantically rich encoders as a measure for image distance in the feature domain.
Although most general vision encoders are trained using natural image cohorts, which vastly differ in appearance and properties from medical images, we demonstrate that the rich embedding spaces could be utilized to drive image registration. An overview of our approach is shown in Fig. \ref{fig:concept}.
To our knowledge, this is the first attempt to investigate whether the features of these modes are applicable as part of distance measures in image registration.

\noindent Our \textbf{main contributions} can be summarized as follows:
\begin{itemize}
    \item Explore whether the features of various general vision encoders are suitable for assessing image distance in deformable image registration
    \item Benchmark the performance of pure feature-based losses vs combined with voxel-intensity-based measures.
    \item Explore the feasibility of integrating feature-based losses in conventional iterative registration.
\end{itemize}

\section{Methods}

Given two $n$-dimensional images, a fixed image \(F\) and a moving image \(M\) with \(F, M:\Omega\subset\mathbbm{R}^n\rightarrow\mathbbm{R}\) and $n\in\{2,3,4\}$, image registration aims to find an optimal spatial transformation $\phi:\mathbbm{R}^n\rightarrow\mathbbm{R}^n$ such that the transformed moving image is most similar to the fixed image. Typically, this is formulated as an optimization problem $\phi^*=\arg \min_{\phi} \mathcal{J}(F, M,\phi)$ where the distance between the images is minimized with constraints on the transformation. We denote the objective function as

\begin{equation}
    \mathcal{J}(F,M,\phi) = \mathcal{D_I}(F,M\circ\phi) + \lambda \mathcal{R}(\phi),
    \label{eq:J}
\end{equation}

\noindent where $\mathcal{D_I}$ is an image-based intensity dissimilarity measure (e.g. MI or NCC) and $\mathcal{R}$, is the regularization on the transformation, which is controlled by the regularization parameter $\lambda$. A common choice is the diffusion regularizer  $\mathcal{R}(\phi)=||\nabla \phi||^2$, used to assure smoothness of the deformation.


\begin{figure}[t]
    \centering
    \includegraphics[width=0.9\textwidth]{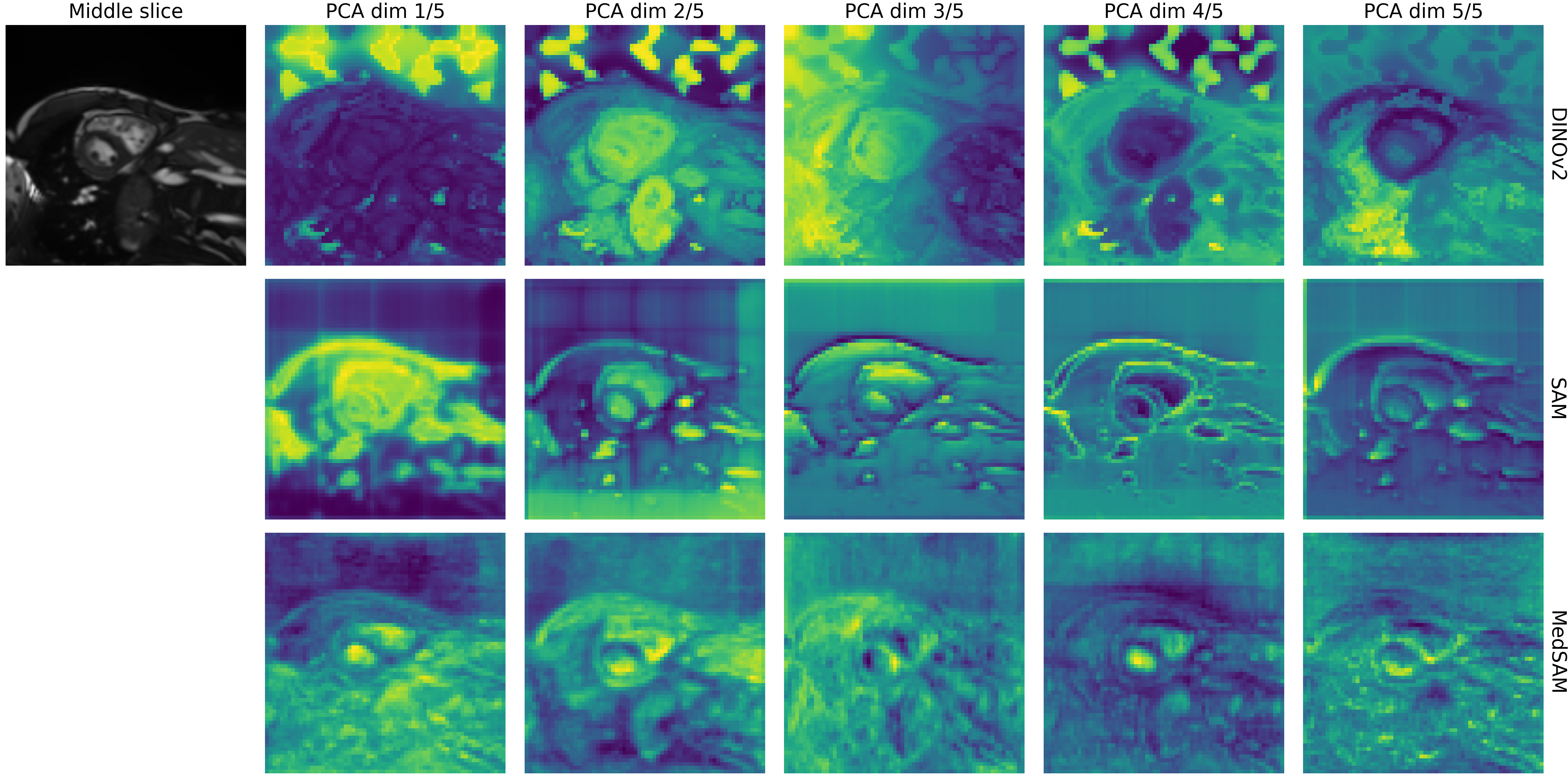}
    \caption{Qualitative comparison of the first 5 principal components of DINOv2, SAM, and MedSAM features for a test image. 
    DINOv2 captures coarser features, while SAM focuses on edges and MedSAM on the underlying texture.}
    \label{fig:features-example}
\end{figure}

In this work, we explore the use of general vision encoder features for medical image registration. In Sec. \ref{sec:feature_extraction}, we describe the encoders used for the extraction of high-dimensional features, and in Sec. \ref{sec:measures}, we present the dissimilarity measures used for feature comparison. Finally, in Sec. \ref{sec:two_variants} we describe two variants of incorporating feature dissimilarity in the registration process.

\subsection{General Vision Encoders for Feature Extraction}\label{sec:feature_extraction}
We compare three different pre-trained encoders: DINOv2 \cite{oquab2023dinov2}, SAM (segment-anything) \cite{kirillov2023segment} and MedSAM (segment-anything in medical images) \cite{medsam}. All three are based on the vision transformer (ViT) architecture \cite{dosovitskiy2020image}. 
DINOv2 was developed using 142M images from 27 classification, segmentation, and depth estimation datasets in a self-supervised manner by self-distillation without labels.
SAM was trained on a large curated segmentation dataset of 11M images. In contrast to DINOv2, SAM employed a supervised learning approach with ground truth segmentations.
While the two models above were trained on natural images, i.e., photographs of real-world objects, the recently published MedSAM model specifically aims to segment medical images. It is a version of SAM fine-tuned with a dataset containing ~1.6M medical images that covers 10 different modalities. Similarly to SAM, it was trained in a supervised manner.

Importantly, none of the three models have been trained on registration tasks. The features of one test image, extracted with the three encoders, are shown in Fig. \ref{fig:features-example}. Since DINOv2 and SAM were trained on 2D RGB images and require such inputs, we use 2D image slices extracted from 3D medical images for all three encoders.
We use these encoders in the registration process by individually mapping the moving and the fixed images to their respective feature spaces. The feature dimension ${N_\mathcal{G}}$ differ between the models - feature maps extracted from DINOv2 have 1024 dimensions, while SAM and MedSAM's have 256 dimensions.

\subsection{Feature-Based Distance Measures}\label{sec:measures}
To use the extracted features $f^F,f^M\in\mathbb{R}^{N_\mathcal{G}}$ of the fixed and warped images for the registration process, we measure their dissimilarity $\mathcal{D_\mathcal{F}}$. Here, $N_\mathcal{G}$ denotes the feature dimension of the respective encoder model. To this end, we explore two different dissimilarity measures: (i) the negative cosine similarity $cos(f^F,f^M) = \frac{f^F\cdot f^M}{||f^F||||f^M||}$ which measures the cosine of the angle between two vectors and which we denote as cosine dissimilarity in the following, and (ii) the mean absolute error (MAE)/L1 distance $L1(f^F, f^M)=\frac{1}{N_\mathcal{G}}\sum_{i}^{N_\mathcal{G}}|f^f_i-f^M_i|$.

\subsection{General Vision Encoder Feature Distance in Image Registration}\label{sec:two_variants}
General vision encoders will be denoted as $\mathcal{G}$ and the extracted features as $f^{I} = \mathcal{G}(I)$ for an image $I$. 
We explore two ways of enhancing the registration objective function with a feature-based distance $\mathcal{D_\mathcal{F}}$:\\

\noindent\textit{Variant 1: Feature distance as the main loss term}\quad Here, we replace the voxel intensity-based image distance measure term $\mathcal{D}(M\circ\phi,F)$ in \ref{eq:J} with a feature-based distance $\mathcal{D_\mathcal{F}}(f^{M\circ\phi},f^F)$:
\begin{equation}
    \mathcal{J}(F,M,\phi) = \mathcal{D_\mathcal{F}}(f^F, f^{M\circ\phi}) + \lambda \mathcal{R}(\phi).
    \label{eq:var1}
\end{equation}

\noindent\textit{Variant 2: Feature distance as additional loss term}\quad Here, rather than exchanging the voxel intensity-based dissimilarity for the feature-based one, we use both terms in the objective function:
\begin{equation}
    \mathcal{J}(F,M,\phi) = (1-\alpha)\mathcal{D_I}(F,M\circ\phi) + \lambda \mathcal{R}(\phi) + \alpha\mathcal{D_\mathcal{F}}(f^F, f^{M\circ\phi}).
    \label{eq:var2}
\end{equation}

The parameter $\alpha\in\mathbb{R}$, with $0\leq \alpha\leq 1$, controls the strength of the feature-based distance term. Note, that $\alpha$ and $\lambda$ have to be chosen carefully.



\section{Experiments and Results}
\subsection{Dataset and Implementation Details}
\subsubsection{Dataset and Data Preprocessing}
We use the ACDC dataset \cite{acdc}, which consists of cardiac MRI images (multi-slice 2-D cine MRI) acquired from 150 patients. Ground truth segmentations of the left ventricular cavity (LV), myocardium (Myo), and right ventricular cavity (RV) are available for both phases. From the 3D volumes, we extract one middle slice per image after cropping around the heart and use the end-diastolic (ED) slices as the fixed and end-systolic (ES) slices as the moving images. The cropped slices have a size of \(128\times128\). Additionally, we resample them to uniform spacing of \(1.8 \, \text{mm} \times 1.8 \, \text{mm} \) and normalize the intensity values to the range $[0,1]$. 

\subsubsection{Evaluation Metrics}
Due to the lack of ground truth deformation fields, we evaluate the registration performance using surrogate measures. 
To assess the registration accuracy, we compare the Dice score (DSC) and the 95th percentile Hausdorff distance (HD95) between the fixed and warped moving segmentation maps.
We consider the class-wise DSC of the three classes LV, RV, Myo, and the mean DSC over those. 
To assess the deformation quality, we evaluate the fraction of negative Jacobians of the displacement field (\%negJdet) that indicates the amount of spatial folding.



\subsubsection{Implementation details}\label{sec:implementaiton-details}
To perform the registration, we use the Airlab~\cite{airlab} registration framework.
This framework not only supports GPU-based computation that accelerates the registration process but also utilizes the Pytorch native autograd for the optimization process.
The latter is particularly useful since it eliminates the need for a manual implementation of the backpropagation function for the encoders.
Additionally, we choose B-Spline free-form deformation~\cite{rueckert1999nonrigid} as a transformation parametrization with a \(24 \times 24\) control point lattice, a learning rate of 5e-4, a regularizer weight of $120$, and 1500 iterations.
The hyperparameters were found empirically by maximizing the DSC while maintaining plausible deformations.
All computations have been performed with Python 3.10 and PyTorch 2.2.2 on an NVIDIA RTX A6000 GPU.
The code is available at \href{https://github.com/compai-lab/2024-miccai-koegl}{github.com/compai-lab/2024-miccai-koegl}.

For a fair comparison, we select openly available encoder instances with approximately similar numbers of trainable parameters, where DINOv2 has $87M$, and SAM and MedSAM have $94M$ parameters. \footnote{For DINOv2, we chose the base model \textit{dinov2\_vitb14\_reg}. 
For SAM and MedSAM, we selected the base models \textit{sam\_vit\_b\_01ec64} and \textit{medsam\_vit\_b}}. 
We repeat the images three times along the channel axis to fit the greyscale images to the three-channel input the encoders require. 
Furthermore, DINOv2 and SAM/MedSAM are trained with different patch sizes, resulting in the resolution of the feature maps being smaller than the input image and differing across the encoders. 
Thus, we upsample the images before extracting the features to obtain a consistent spatial resolution for the feature maps across all encoders. The feature maps have a spatial resolution of $64\times 64$. 
Note that the MedSAM model can handle both 2D and 3D images. However, since the other encoders require 2D inputs, we restrict ourselves to 2D images. 



\subsection{Results}
\subsubsection{Comparison of Feature-Based Distance Measures' Curves for Rotations and Translations}
In this experiment, we investigate the effectivity of the L1 and cosine dissimilarity measures between features when recovering rigid transformations.  
We conduct two experiments, one where we rotate the image by steps of $\frac{\pi}{32}rad$ in the range of $[-90^\circ,90^\circ]$ and another one where we translate the image along one axis by steps of $1.2mm$ in the range [-57.6, 57.6] mm.
To this end, we choose one test image and gradually transform it. 
At each step, we encode the transformed and original images, normalize the features to the range $[0, 1]$, and measure the dissimilarity between the two feature maps.

\begin{figure}[t]
    \centering
    \includegraphics[width=\textwidth]{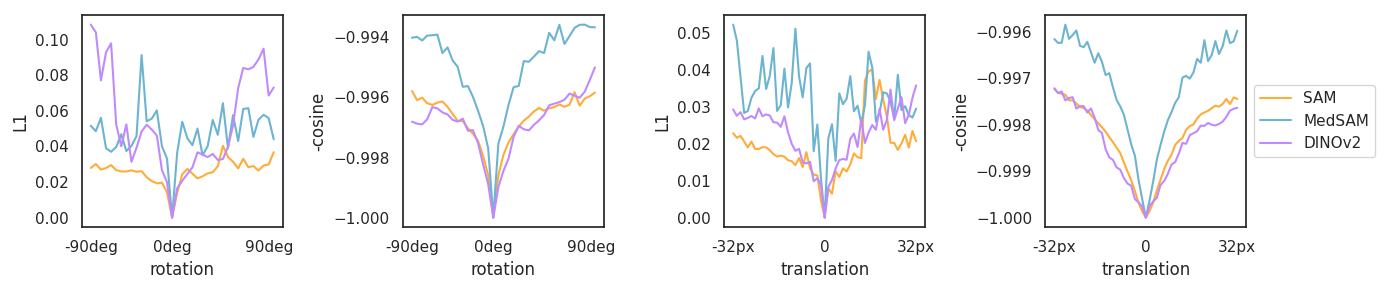}
    \caption{Comparison of the L1 and cosine dissimilarity measures on the features of the three encoder models for rotation and translation of one test image.
    For both rotation and translation tasks, the cosine dissimilarity curves appear smoother and with a wider capture range making it a better choice as a dissimilarity measure.}
    \label{fig:rot_trans}
\end{figure}

Figure \ref{fig:rot_trans} shows the curves for the measured feature distances with L1 and cosine dissimilarity for all three encoders for rotation and translation. 
In all cases, there is a clear global minimum at zero transformation, which is a desired behavior for a dissimilarity measure in the context of registration. 
Moreover, the cosine dissimilarity curves demonstrate a wider capture range for both rotation and translation while being smoother than the L1 curves.
The latter appear less steep and not so smooth, demonstrating more local minima, making them more difficult to optimize. 
%

\subsubsection{Feature-Based Distance Measures in Medical Image Registration}
In this experiment, we investigate whether the general vision encoders' features can effectively be used within a distance measure for image registration. 
To evaluate this we conduct three comparisons: (i) using the feature distance as the sole dissimilarity measure (\textit{variant 1}) versus using it as additional guidance alongside the image intensity-based distance (\textit{variant 2}), (ii) using the L1 versus the cosine dissimilarity as distance measures between features, and (iii) using each of the three general vision encoders. 
As the baseline, we use the standard registration objective of Eq. \ref{eq:J} with NCC as an image dissimilarity measure and a weighted diffusion regularization term. 

\begin{figure}[t]
    \centering
    \includegraphics[width=0.9\textwidth]{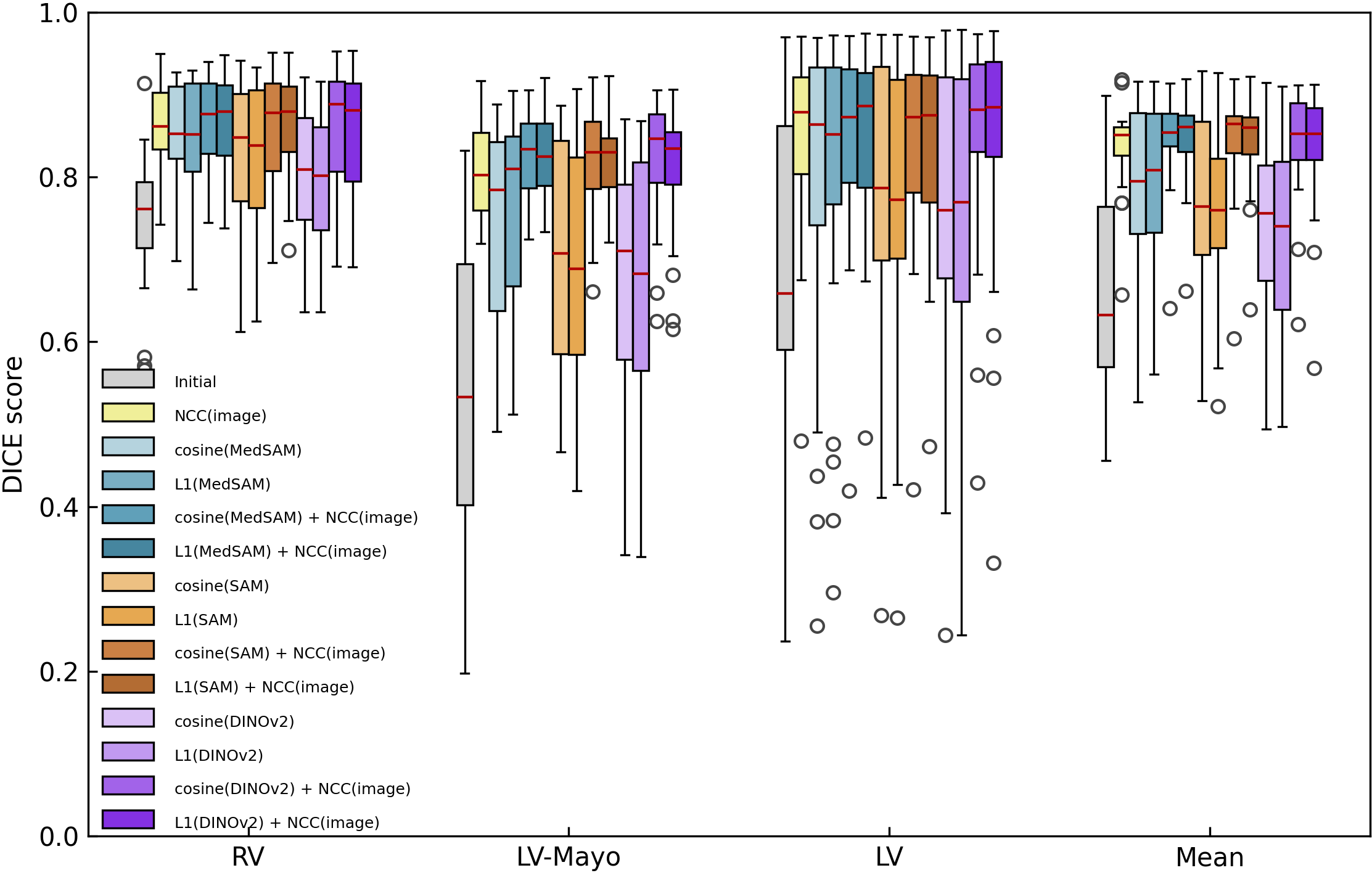}
    \caption{Quantitative comparison: Class-wise and mean DSC scores before and after registration. Methods using the same encoder share the same color with varying hues. The boxplots are grouped horizontally by class: right ventricular cavity (RV), left ventricular myocardium (LV-Mayo), left ventricular cavity (LV), and the mean over all classes. We see that \textit{variant 2} outperforms the baseline in both class-wise and mean DSC.}
    \label{fig:quantitative-results}
\end{figure}

\textbf{Variant 1 vs. variant 2}:
Tab.~\ref{tab:quantitavie_results} and Fig. \ref{fig:quantitative-results} demonstrate that incorporating general vision encoded features as an auxiliary dissimilarity term (\textit{variant 2}) enhances registration performance compared to the baseline.
This improvement is observed for both dissimilarity measures and all encoder models, as indicated by the mean DSC and mean HD95.
Specifically, a mean DSC of $0.845$ and mean HD95 of $1.765$ is achieved using the MedSAM encoder with cosine dissimilarity, surpassing the baseline mean DSC of $0.836$ and mean HD95 of $1.916$. 
This trend is also evident in class-wise DSC (see Fig. \ref{fig:quantitative-results}).
On the other hand, \textit{variant 1}, which does not utilize encoded features, results in decreased registration performance compared to the baseline.

\textbf{L1 distance vs. cosine dissimilarity}:
Firstly, the cosine dissimilarity generally improves the HD95 across most experiments for both variants.
For \textit{variant 2}, it consistently achieves a slightly higher mean DSC compared to the L1 distance, as shown in \ref{tab:quantitavie_results}. 
Conversely, for \textit{variant 1}, this trend holds true for two out of three encoder models (DINOv2 and SAM). 

\begin{table}[t]
\centering
\begin{scriptsize}
    \caption{Quantitavie results: class-wise and mean DSC, HD95 and sdlogJ before and after registration. We see that \textit{variant 2} consistently outperforms the baseline in terms of DSC and HD95.}
    \begin{tabular}{ccccc}
    \toprule\toprule
        \textbf{Encoder $\mathcal{G}$} & \textbf{Variant} & \textbf{DSC\( \uparrow \)} & \textbf{HD95\( \downarrow \)} & \textbf{sdlogJ\( \downarrow \)} \\
        \toprule
          no encoder & initial & 0.651 $\pm$ 0.127 & 4.125 $\pm$ 1.816 & - \\
          \midrule
          no encoder & NCC & 0.836 $\pm$ 0.051 & 1.916 $\pm$ 1.154 & 0.385 $\pm$ 0.111\\
          \midrule
          \multirow{4}{*}{DINOv2} & L1 & 0.730 $\pm$ 0.113 & 3.095 $\pm$ 1.518 & 0.264 $\pm$ 0.058\\
           & Cos & 0.742 $\pm$ 0.108 & 2.936 $\pm$ 1.465 & 0.244 $\pm$ 0.053 \\
           & L1 + NCC & 0.833 $\pm$ 0.075 & 1.859 $\pm$ 1.127 & 0.330 $\pm$ 0.087 \\
           & Cos + NCC & 0.844 $\pm$ 0.063 & \textbf{1.711 $\pm$ 0.918} & 0.321 $\pm$ 0.082 \\
           \midrule
          \multirow{4}{*}{SAM} & L1 & 0.758 $\pm$ 0.106 & 3.052 $\pm$ 1.579 & \textbf{0.161 $\pm$ 0.030} \\
           & Cos & 0.761 $\pm$ 0.106 & 2.980 $\pm$ 1.614 & 0.226 $\pm$ 0.040 \\
           & L1 + NCC & 0.841 $\pm$ 0.059 & 1.862 $\pm$ 1.184 & 0.331 $\pm$ 0.100 \\
           & Cos + NCC & 0.842 $\pm$ 0.063 & 1.900 $\pm$ 1.204 & 0.348 $\pm$ 0.095 \\
           \midrule
          \multirow{4}{*}{MedSAM} & L1 & 0.798 $\pm$ 0.100 & 2.363 $\pm$ 1.422 & 0.243 $\pm$ 0.055 \\
           & Cos & 0.791 $\pm$ 0.105 & 2.534 $\pm$ 1.572 & 0.333 $\pm$ 0.061 \\
           & L1 + NCC & 0.845 $\pm$ 0.055 & 1.819 $\pm$ 1.133 & 0.363 $\pm$ 0.104 \\
           & Cos + NCC & \textbf{0.845 $\pm$ 0.054} & 1.765 $\pm$ 1.077 & 0.395 $\pm$ 0.096 \\
         \bottomrule\bottomrule
    \end{tabular}
    \label{tab:quantitavie_results}
    \end{scriptsize}
\end{table}

\textbf{DINOv2 vs. SAM vs. MedSAM}:
The results indicate that MedSAM outperforms SAM in terms of both DSC and HD95, particularly in \textit{variant 1}. 
In \textit{variant 2}, the DSC and HD95 show less variation between the models compared to \textit{variant 1}. 
Here, MedSAM achieves the best DSC while DINOv2 achieves the best HD95. 
Last but not least, SAM achieves the smoothest deformations with \textit{variant 1}. 
Finally, there is no observable folding across all models and variants.

\begin{figure}[t]
    \raggedright

    \begin{subfigure}[t]{0.24\textwidth}
        \centering
        \includegraphics[height=2.3cm]{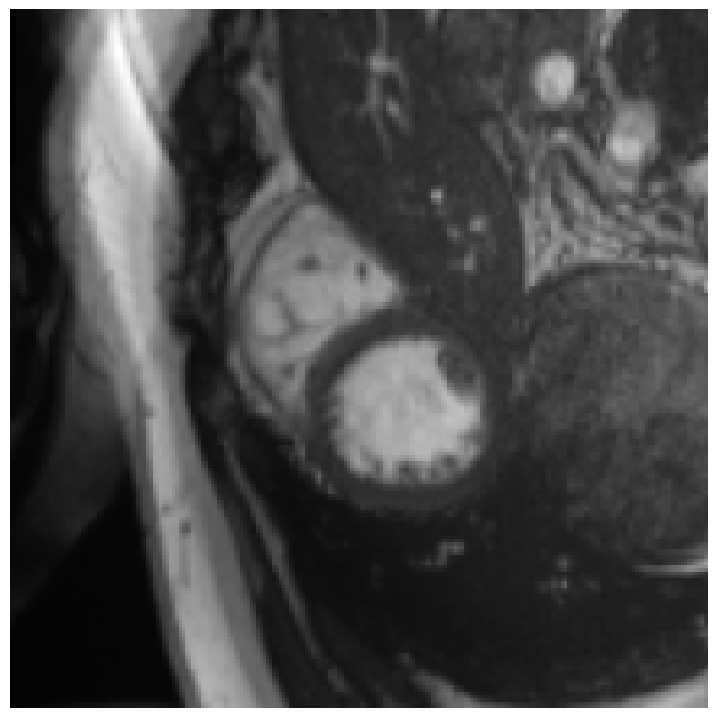}
        \caption*{\small Fixed image}
    \end{subfigure}
    \begin{subfigure}[t]{0.24\textwidth}
        \centering
        \includegraphics[height=2.3cm]{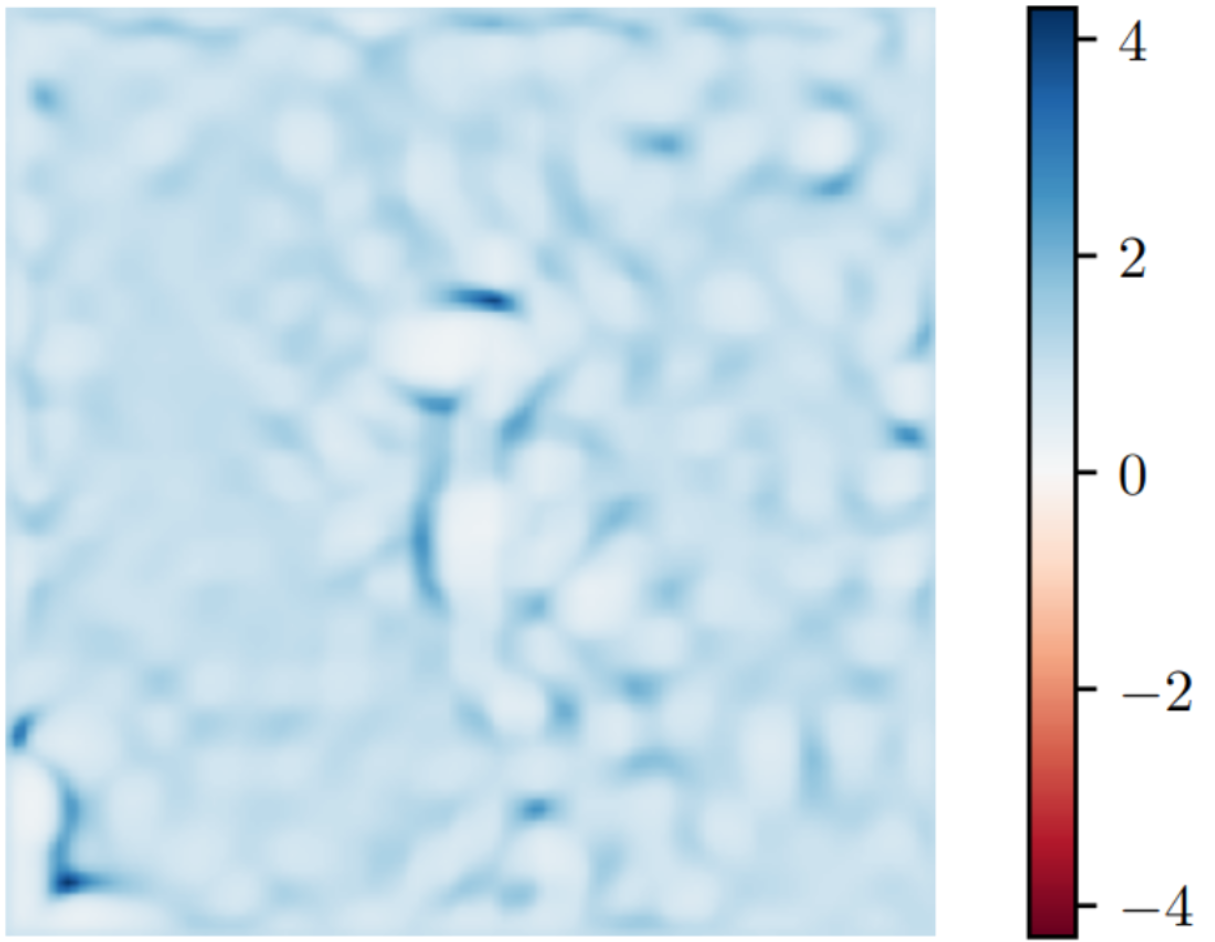}
        \caption*{\small DINOv2}
    \end{subfigure}
    \begin{subfigure}[t]{0.24\textwidth}
        \centering
        \includegraphics[height=2.3cm]{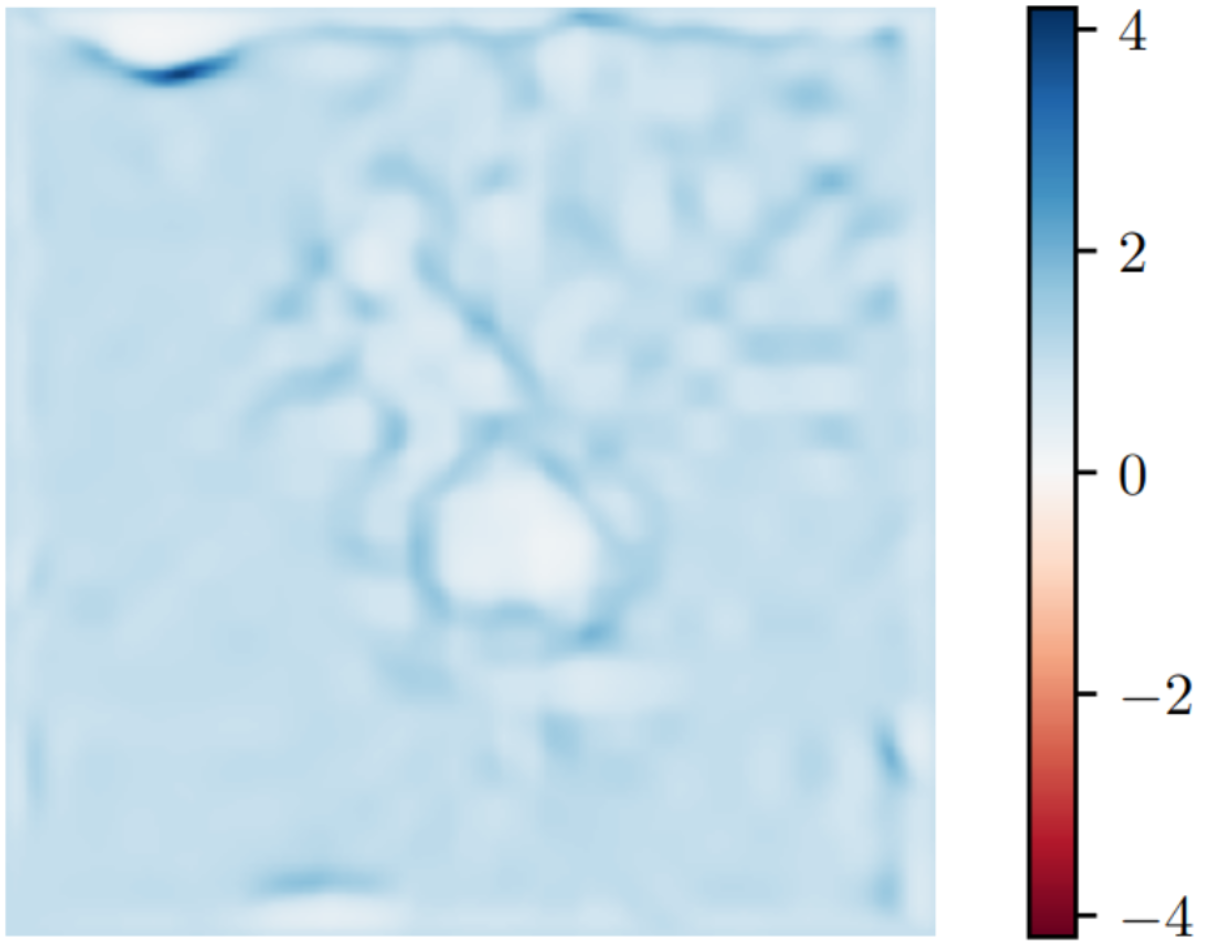}
        \caption*{\small SAM}
    \end{subfigure}
    \begin{subfigure}[t]{0.24\textwidth}
        \centering
        \includegraphics[height=2.3cm]{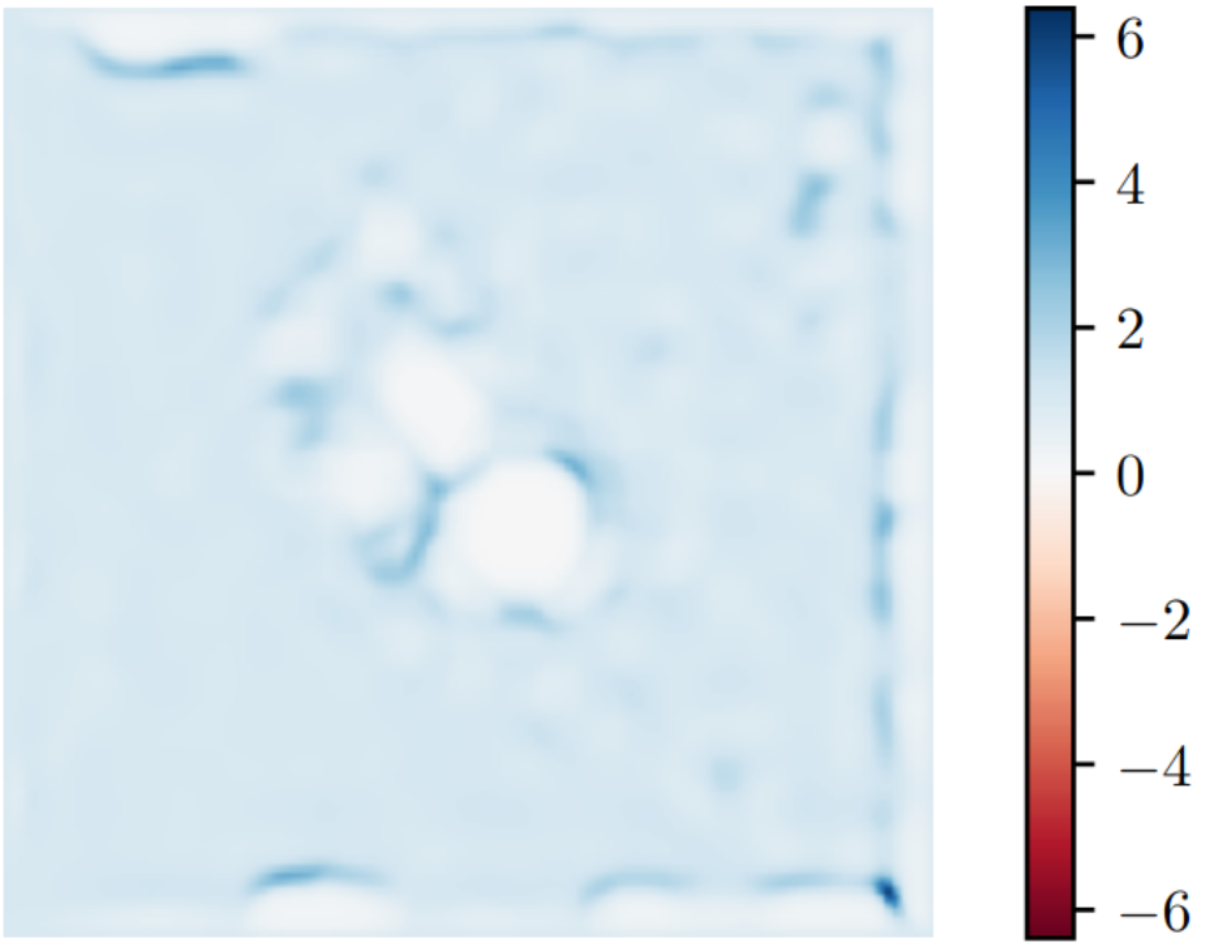}
        \caption*{\small NCC}
    \end{subfigure}
    \caption{Qualitative comparison of the Jacobian determinant maps from \textit{variant 1} with the DINOv2 and SAM encoders, as well as the baseline. From left to right, we see that the Jacobian determinant map of DINOv2 has many local deformations in the background regions, while the one of SAM has fewer. The baseline transformation appears smoother outside of the heart. The fixed and moving images differ only in the heart, so we would expect most of the deformation to occur there. White areas show shrinkage, and blue areas expansion.}
    \label{fig:jacdet-comparison}
\end{figure}

\newcommand{\myfeatwidth}{0.17}
\begin{figure}[t]
    \centering
    \begin{subfigure}[t]{\myfeatwidth\textwidth}
        \centering
        \includegraphics[width=\textwidth]{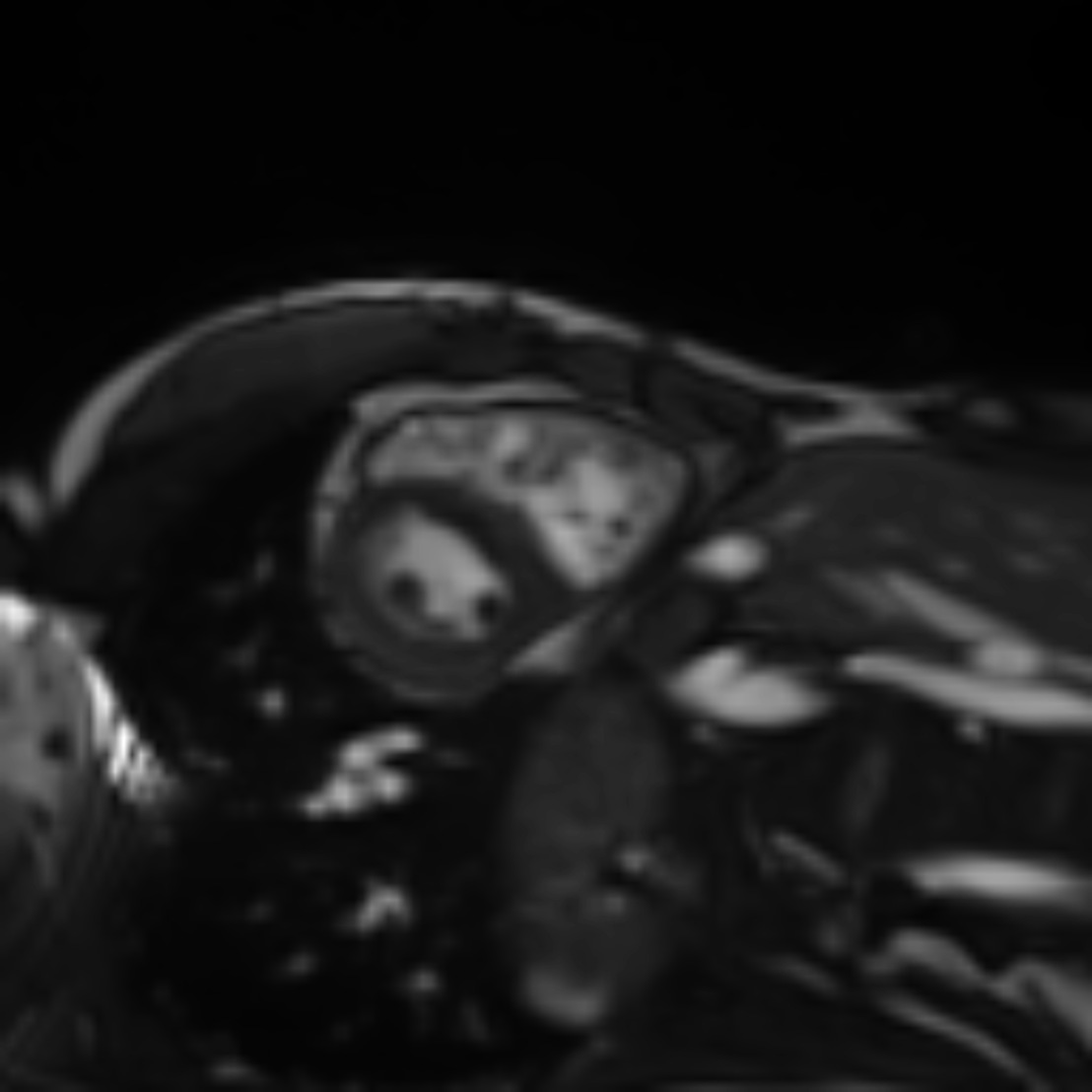}
        \caption*{\tiny Middle slice}
    \end{subfigure}
    \begin{subfigure}[t]{\myfeatwidth\textwidth}
        \centering
        \includegraphics[width=\textwidth]{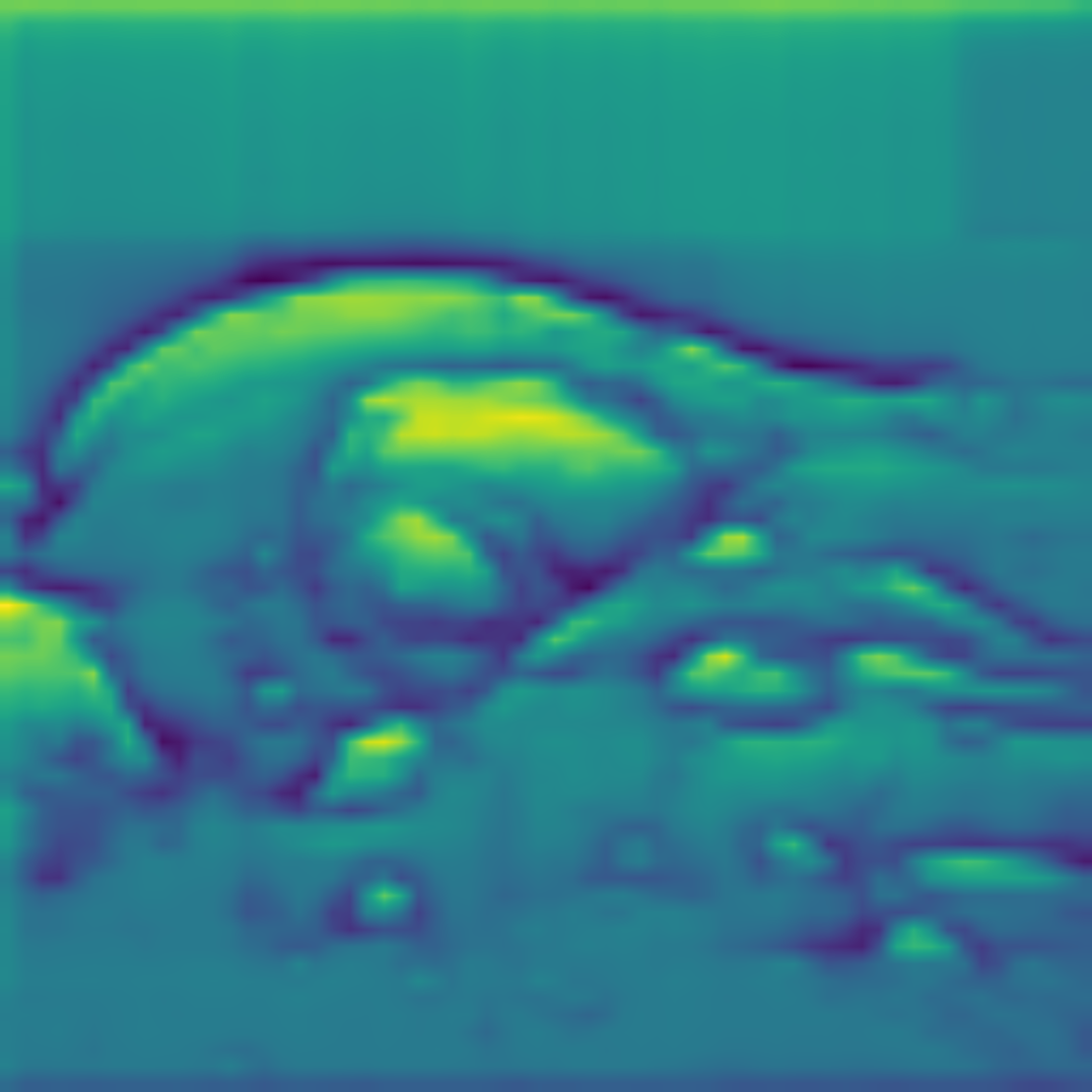}
        \caption*{\tiny Middle slice feature}
    \end{subfigure}
    \begin{subfigure}[t]{\myfeatwidth\textwidth}
        \centering
        \includegraphics[width=\textwidth]{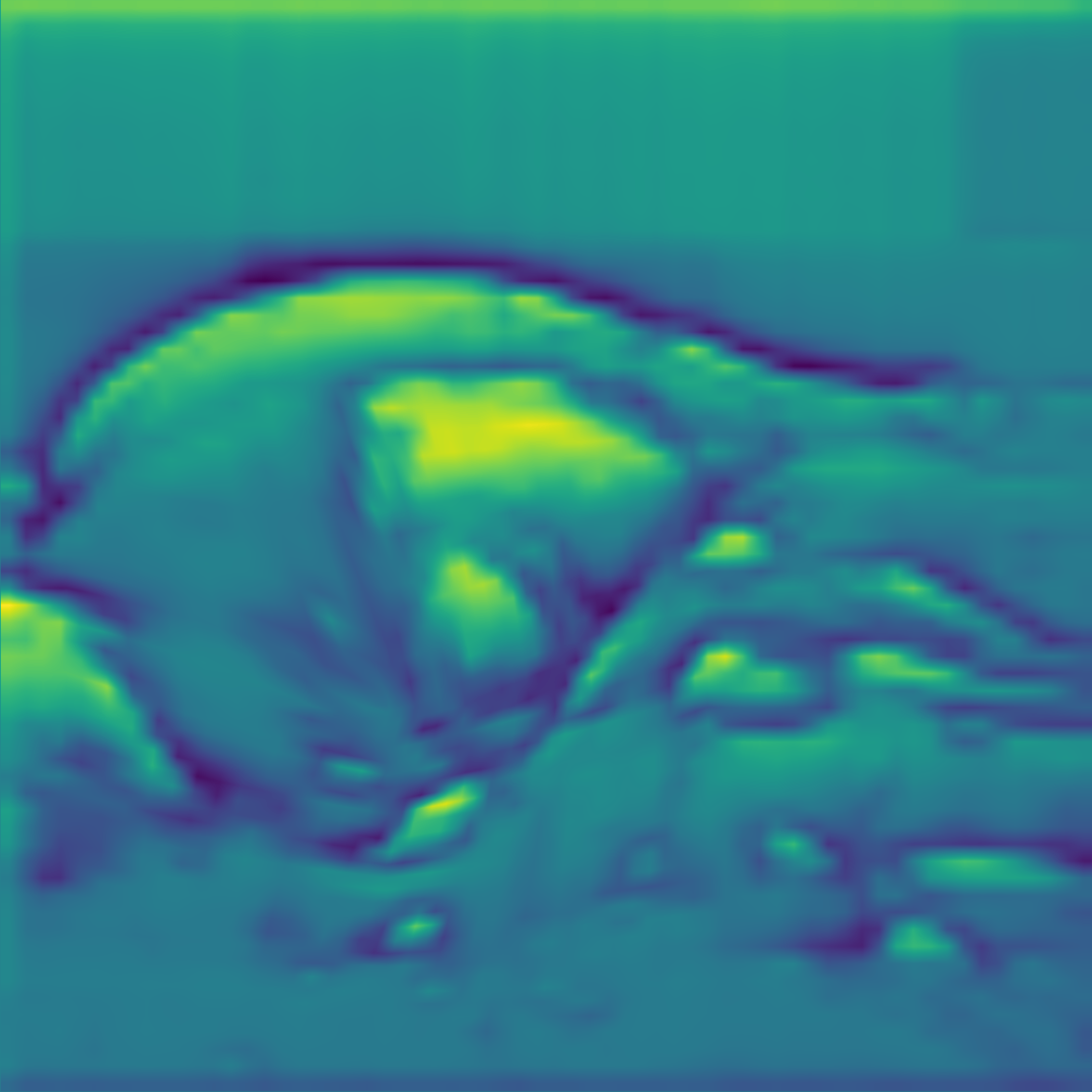}
        \caption*{\tiny encode $\rightarrow$ \textnormal{warp}}
    \end{subfigure}
    \begin{subfigure}[t]{\myfeatwidth\textwidth}
        \centering
        \includegraphics[width=\textwidth]{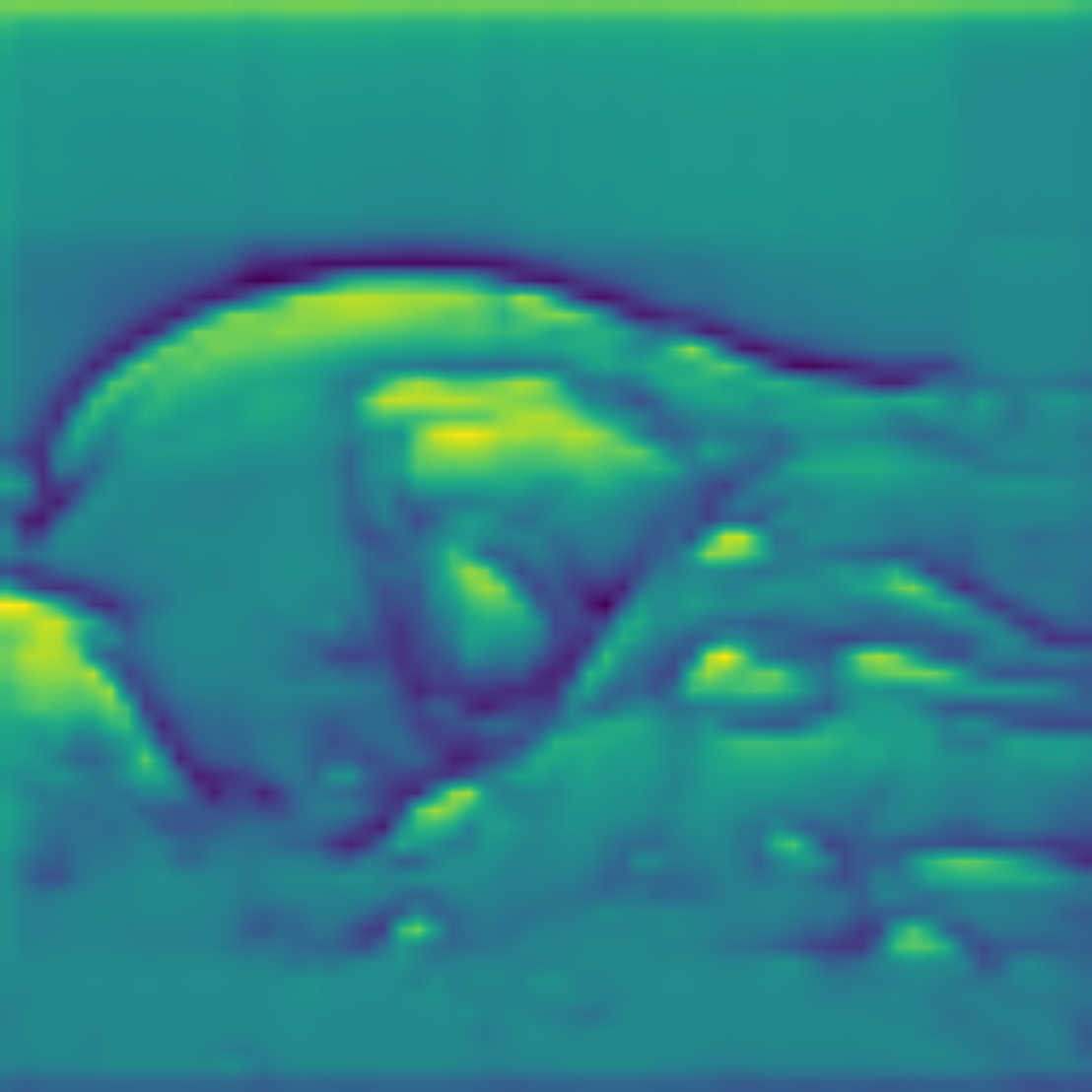}
        \caption*{\tiny warp $\rightarrow$ \textnormal{encode} }
    \end{subfigure}
    \begin{subfigure}[t]{\myfeatwidth\textwidth}
        \centering
        \includegraphics[width=\textwidth]{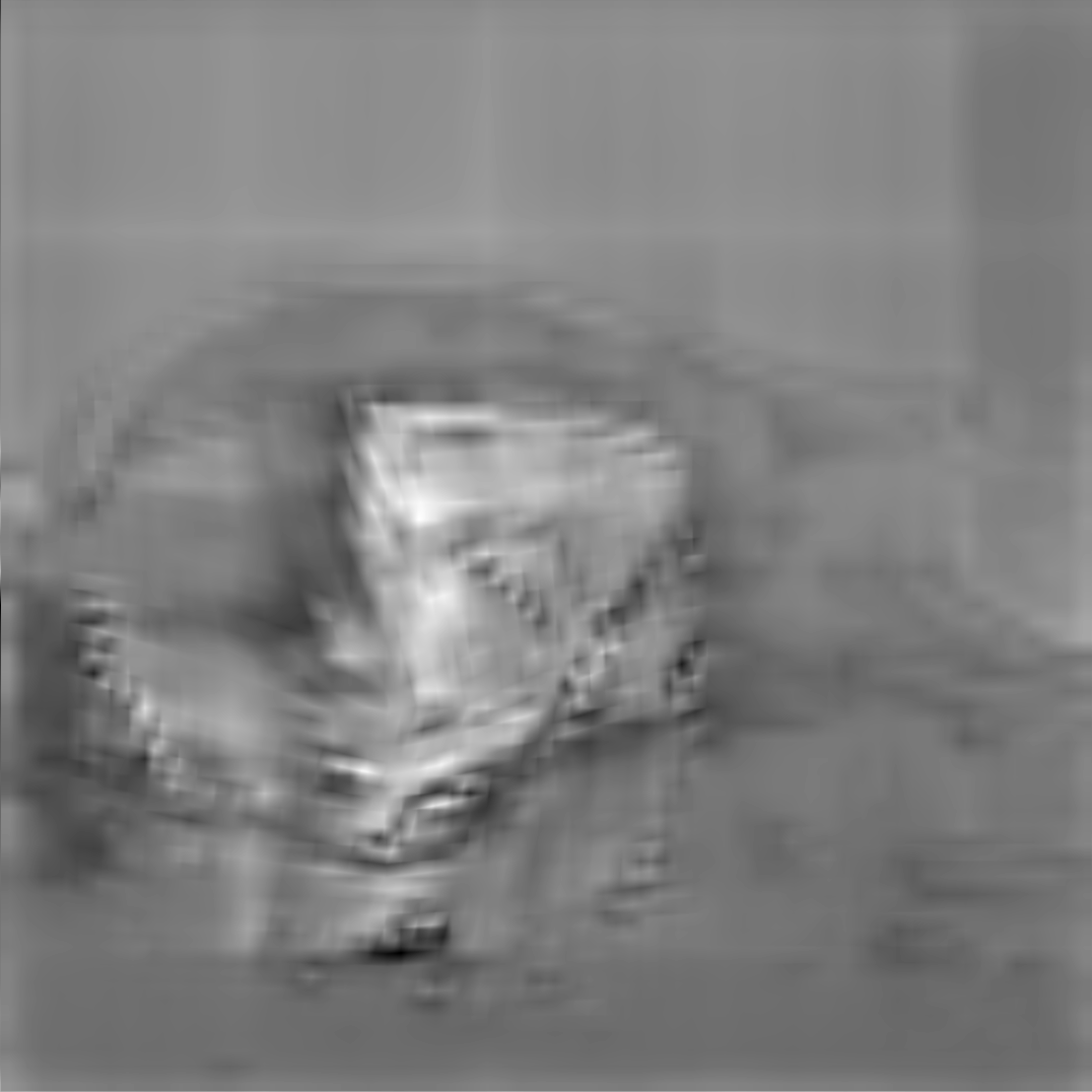}
        \caption*{\tiny Difference image}
    \end{subfigure}
        \caption{We find that feature extraction and warping are not commutative. From left to right, the figure shows the original image, the encoded image, the warped features, the encoded warped image, and the difference image between the latter two. We observe that the obtained features differ.}
    \label{fig:commutative-features}
\end{figure}

As expected, the MedSAM encoder achieves better results than SAM since it was fine-tuned on medical data and thus should be able to capture anatomical structures better than the general vision encoders.  
Nevertheless, it is surprising that the DINOv2 encoder, despite not seeing medical images during training, improves the results for \textit{variant 2}. 
A reason that its features achieve such low performance when used in \textit{variant 1} could be connected to the feature artifacts in the background regions (see Fig. \ref{fig:features-example}. 
This hypothesis is supported by Fig. \ref{fig:jacdet-comparison}: It shows the three Jacobian determinant maps that have been obtained with \textit{variant 1} for the DINOv2 and SAM encoders as well as the baseline registration. 
In the DINOv2-based deformation, local deformations are also visible in the background, while for the NCC-based registration, the deformation concentrates around the relevant regions of the heart. 
The general lack of folding in the qualitative results and Fig. \ref{fig:jacdet-comparison} is most likely achieved with the relatively high regularization weight of the diffusion regularizer.
Additionally, we find that the encoding of images and warping of features are not commutative: changing the order of warping and encoding leads to different features (see Fig. \ref{fig:commutative-features}).

Overall, the fact that \textit{variant 1} is outperformed by the baseline, which in turn is outperformed by \textit{variant 2} on the one hand reveals that neither the general vision encoders nor the fine-tuned MedSAM encoder is suitable as the main driver of the registration optimization. 
An explanation for this could be that the feature maps in the experiments are only half the size of the original images. Thus, some spatial information is lost, which the high feature dimensionality cannot compensate for. 
On the other hand, this also shows that feature-based dissimilarity measures can successfully serve as additional guidance in medical image registration, even if the encoder models have not explicitly been trained on medical data, as is the case with the DINOv2 encoder. 

\newcommand{\myheight}{2.75cm}
\newcommand{\mywidth}{0.2}
\newcommand{\myhorizspaceing}{2.5mm}
\newcommand{\captionvspace}{-7mm}

\begin{figure}[]
    \raggedright
    \begin{minipage}[b]{\mywidth\textwidth}
        \centering
        \includegraphics[height=\myheight]{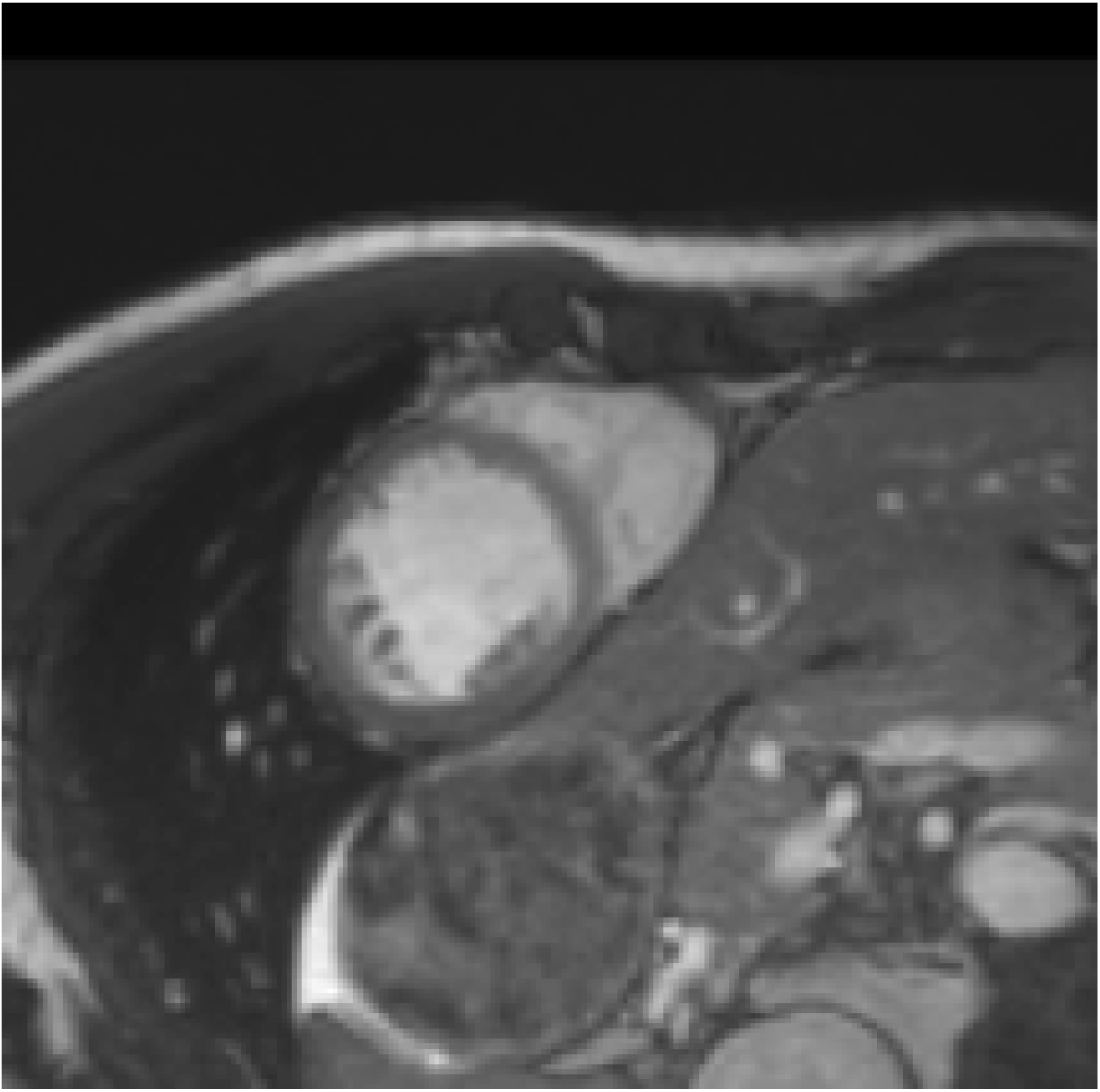}
        \vspace{\captionvspace} 
        \caption*{\scriptsize Moving}
    \end{minipage}
    \hspace{\myhorizspaceing}
    \begin{minipage}[b]{\mywidth\textwidth}
        \centering
        \includegraphics[height=\myheight]{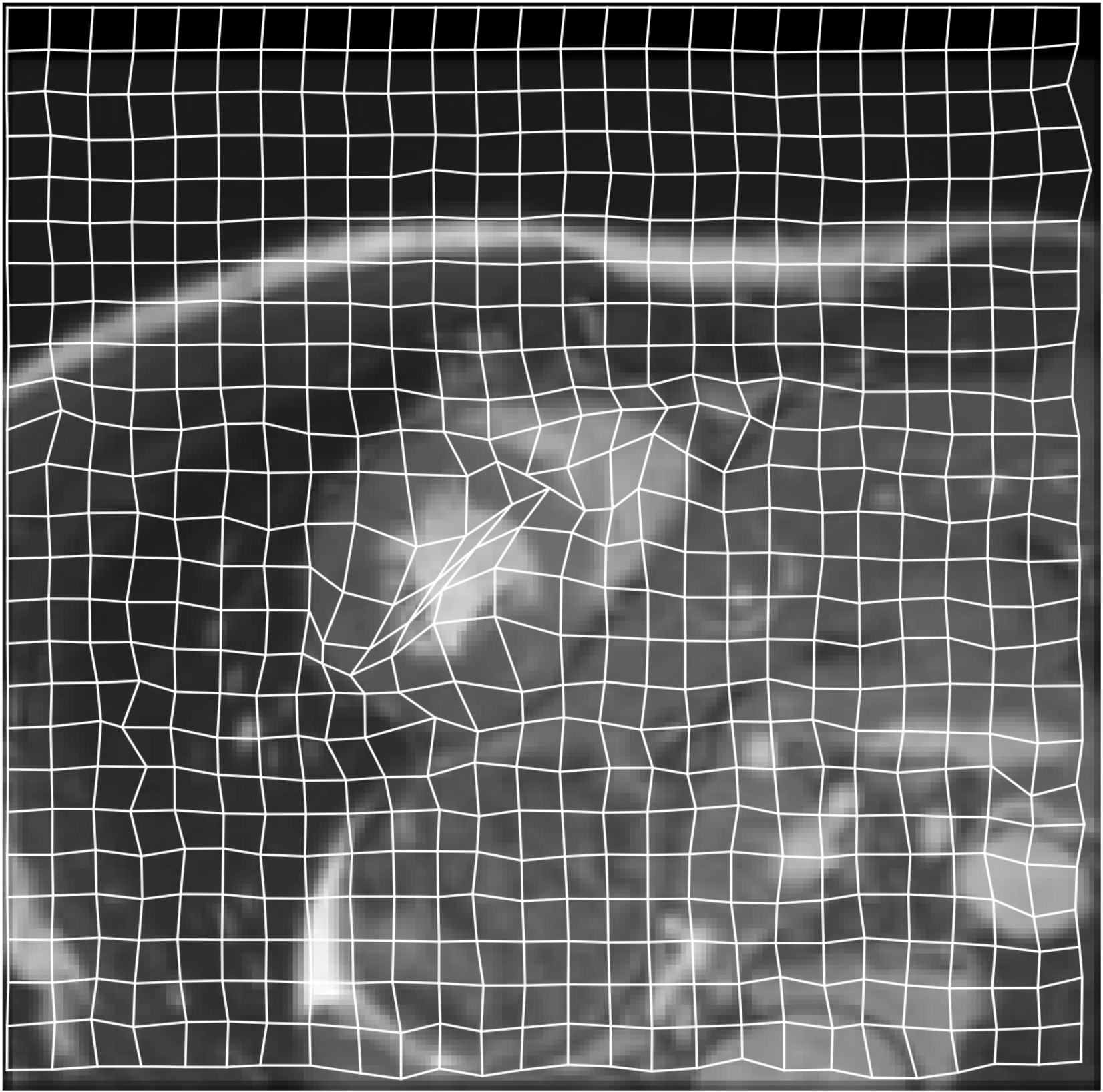}
        \vspace{\captionvspace} 
        \caption*{\scriptsize Deformation field}
    \end{minipage}
    \hspace{\myhorizspaceing}
    \begin{minipage}[b]{\mywidth\textwidth}
        \centering
        \includegraphics[height=\myheight]{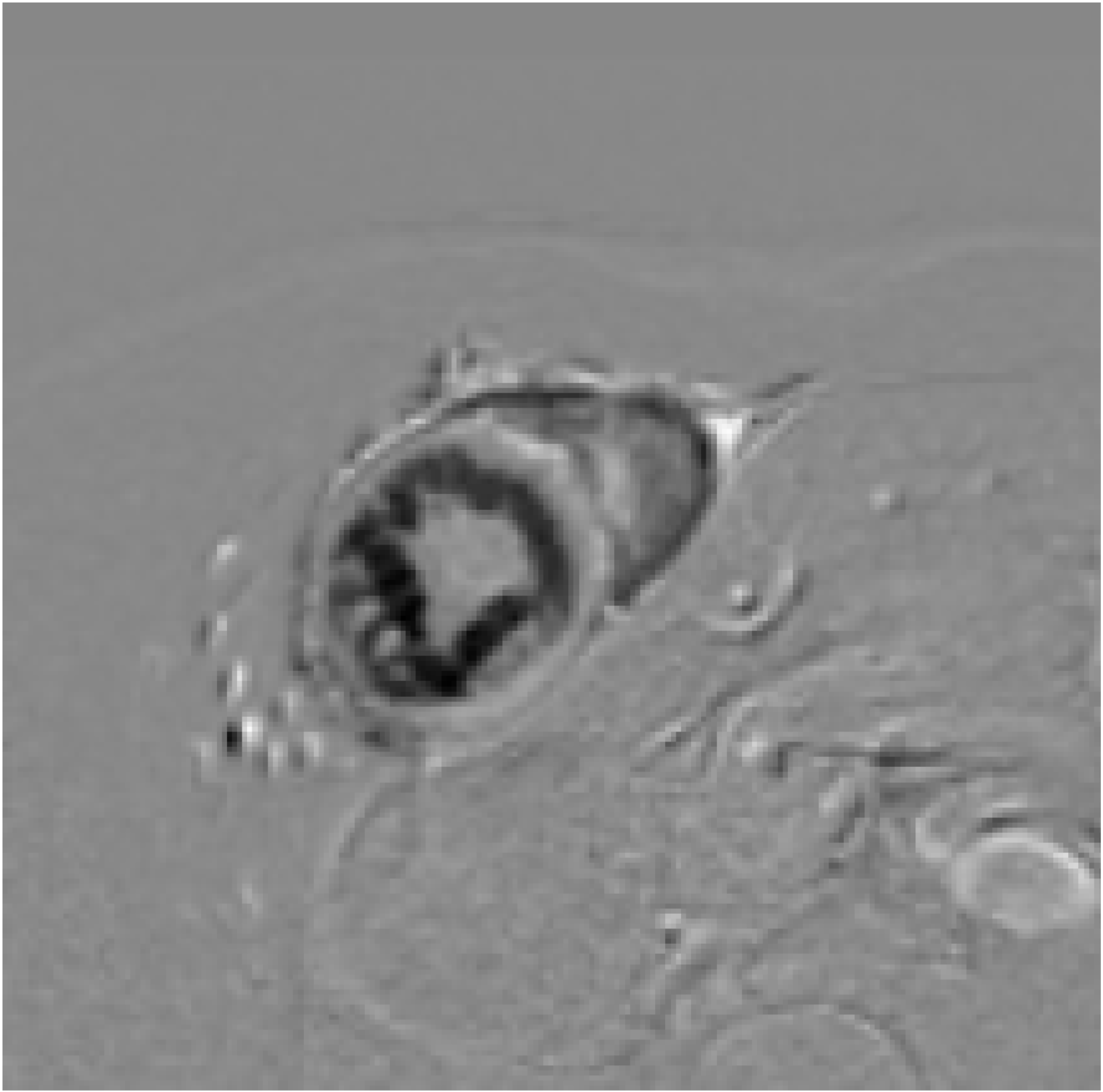}
        \vspace{\captionvspace} 
        \caption*{\scriptsize Diff. before}
    \end{minipage}
    \hspace{\myhorizspaceing}
    \begin{minipage}[b]{\mywidth\textwidth}
        \centering
        \includegraphics[height=\myheight]{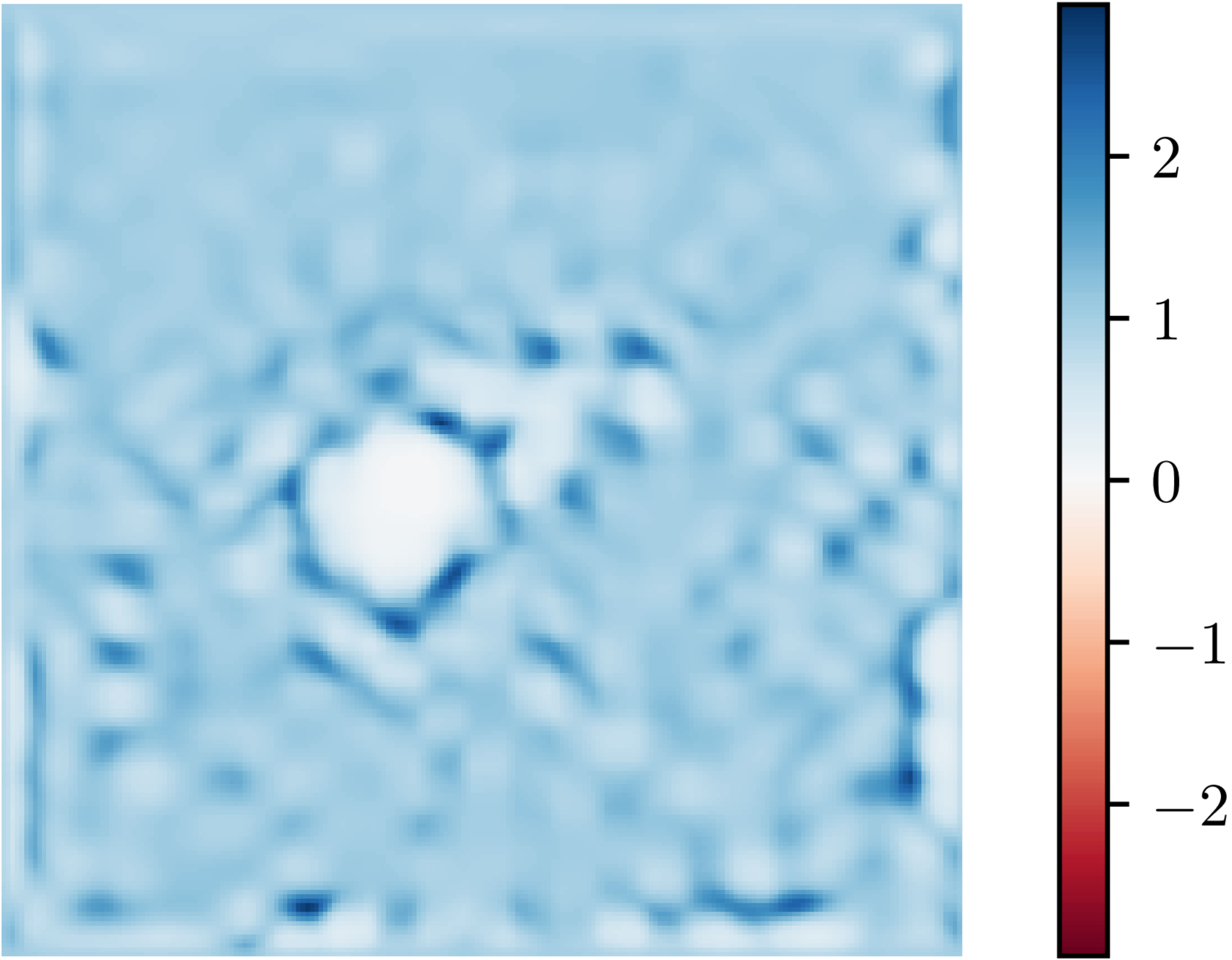}
        \vspace{\captionvspace} 
        \caption*{\scriptsize Jacobian det.}
    \end{minipage}
    
    \vspace{2mm}
    
    \begin{minipage}[b]{\mywidth\textwidth}
        \centering
        \includegraphics[height=\myheight]{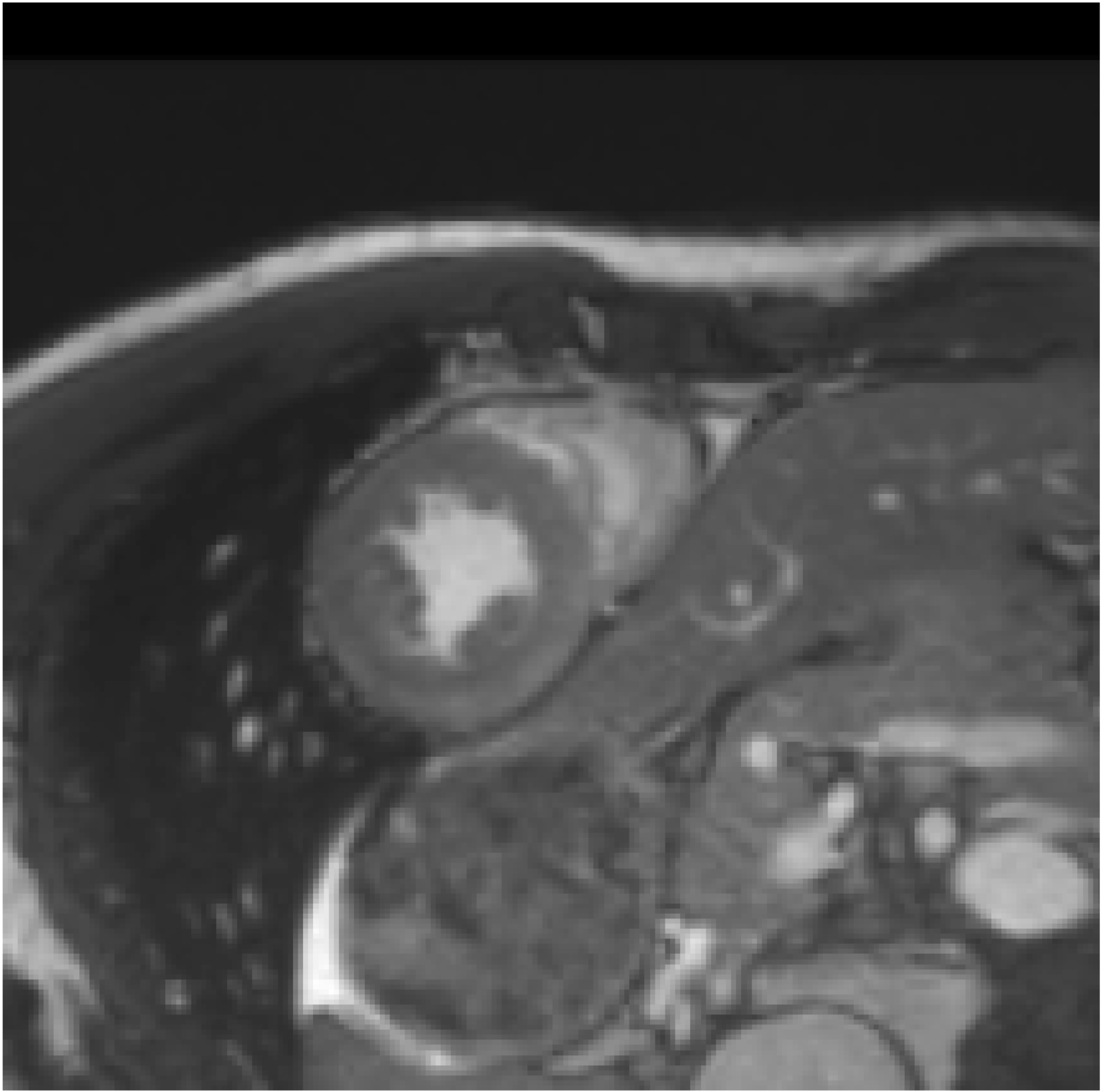}
        \vspace{\captionvspace} 
        \caption*{\scriptsize Fixed}
    \end{minipage}
    \hspace{\myhorizspaceing}
    \begin{minipage}[b]{\mywidth\textwidth}
        \centering
        \includegraphics[height=\myheight]{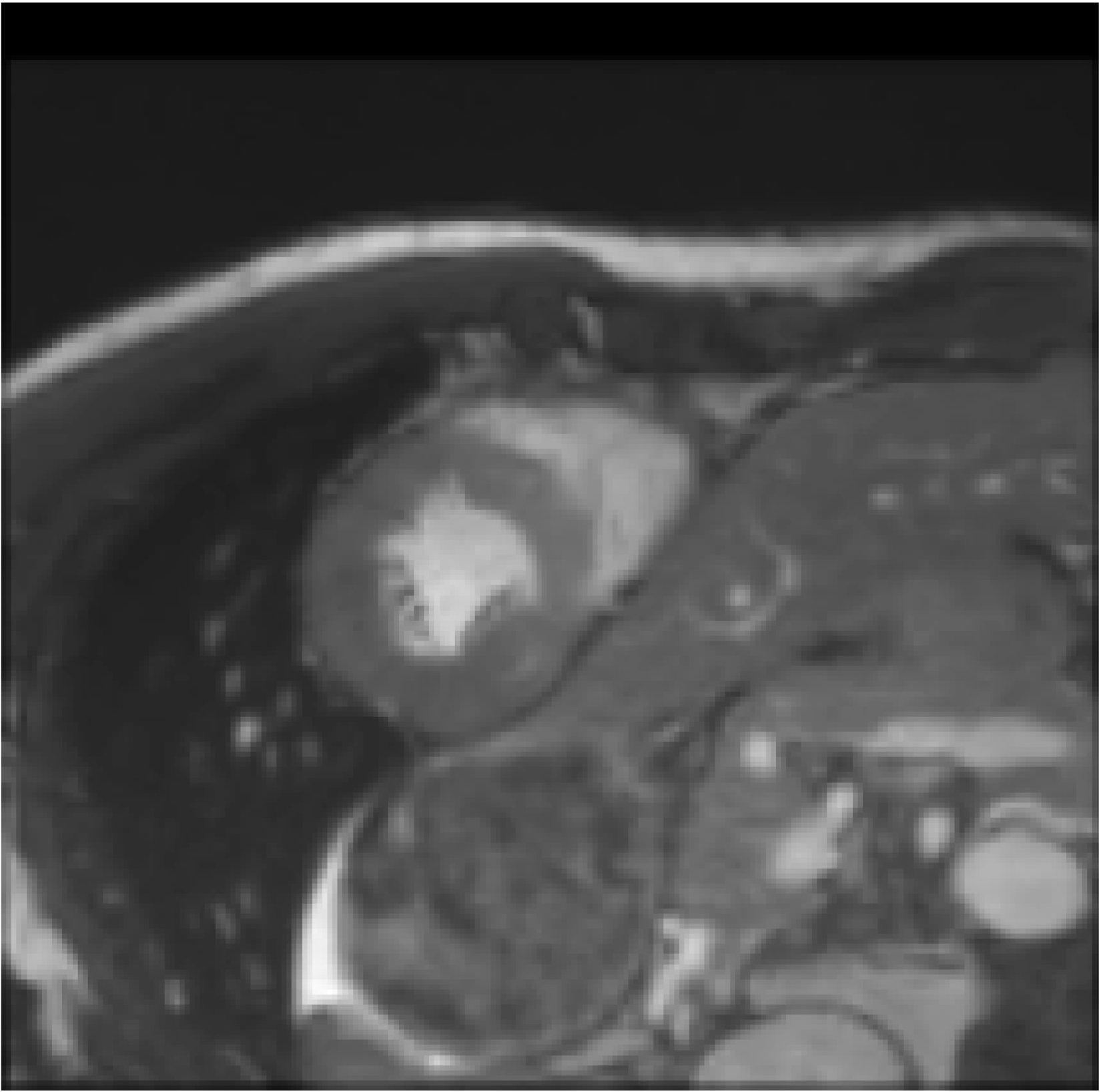}
        \vspace{\captionvspace} 
        \caption*{\scriptsize Warped}
    \end{minipage}
    \hspace{\myhorizspaceing}
    \begin{minipage}[b]{\mywidth\textwidth}
        \centering
        \includegraphics[height=\myheight]{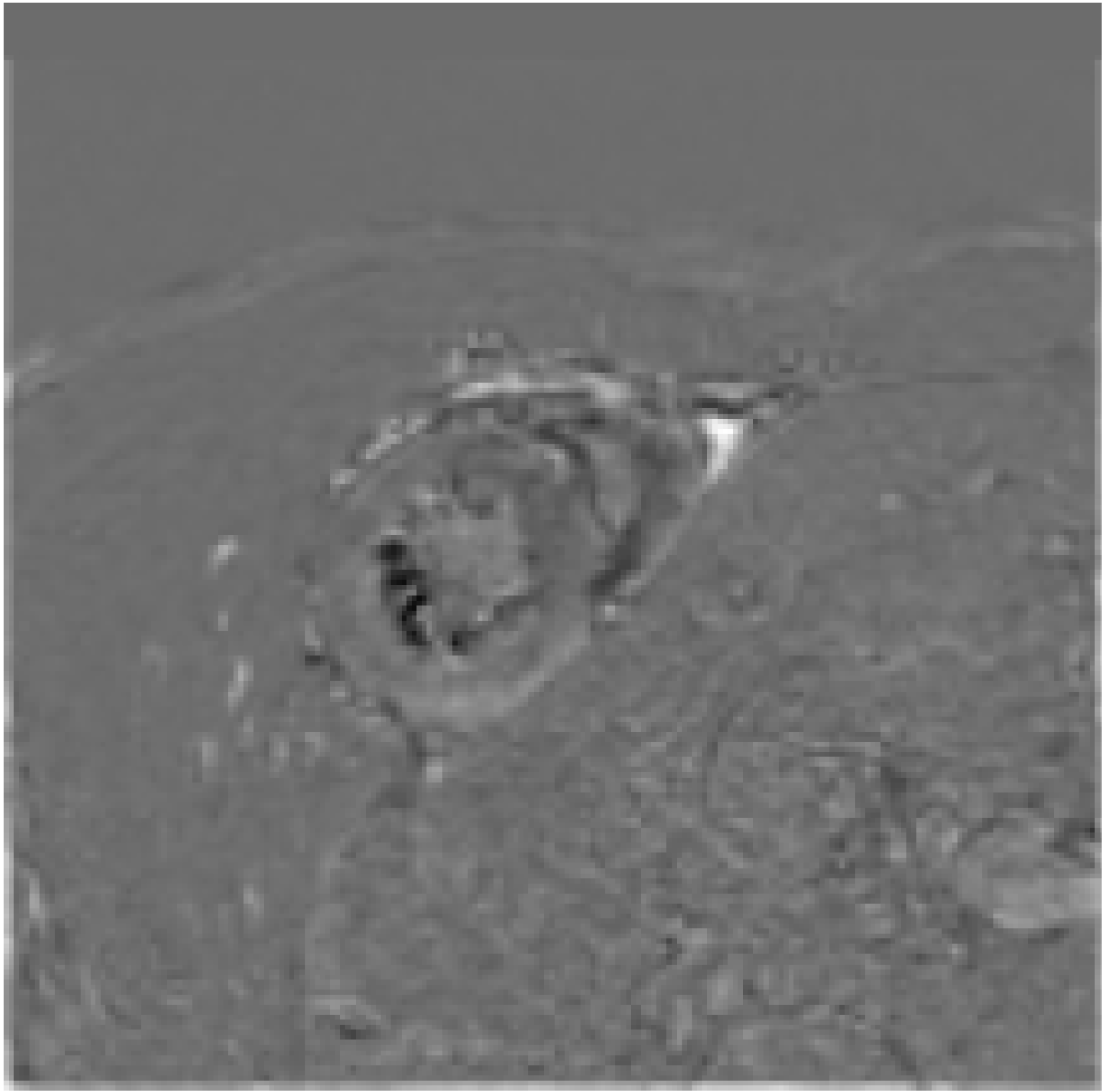}
        \vspace{\captionvspace} 
        \caption*{\scriptsize Diff. after}
    \end{minipage}
    \hspace{\myhorizspaceing}
    \begin{minipage}[b]{\mywidth\textwidth}
        \centering
        \includegraphics[height=\myheight]{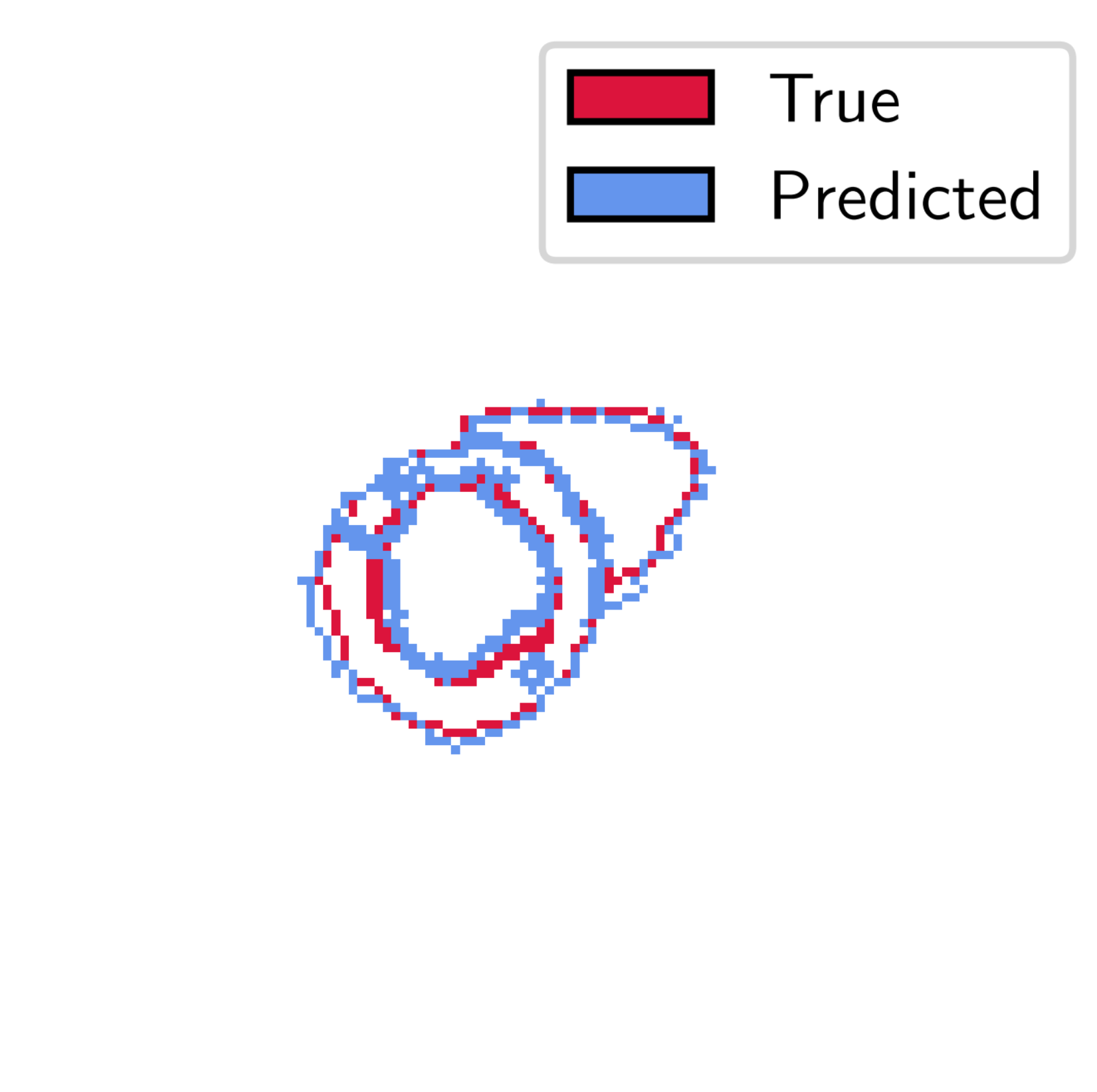}
        \vspace{\captionvspace} 
        \caption*{\scriptsize Segmentations}
    \end{minipage}
    \caption{Qualitative registration results with \textit{variant 2} with $\alpha = 0.3$ for one image pair. The difference images before and after show a clear improvement and the Jacobian determinant map shows shrinkage inside the ventricles, as expected. The segmentation overlap improves from a DSC of 0.622 and HD95 of 7.13 to 0.860 and 1.28, respectively.}
    \label{fig:example-qualitative}
\end{figure}
\begin{figure}[t]
    \centering
    \includegraphics[width=0.7\textwidth]{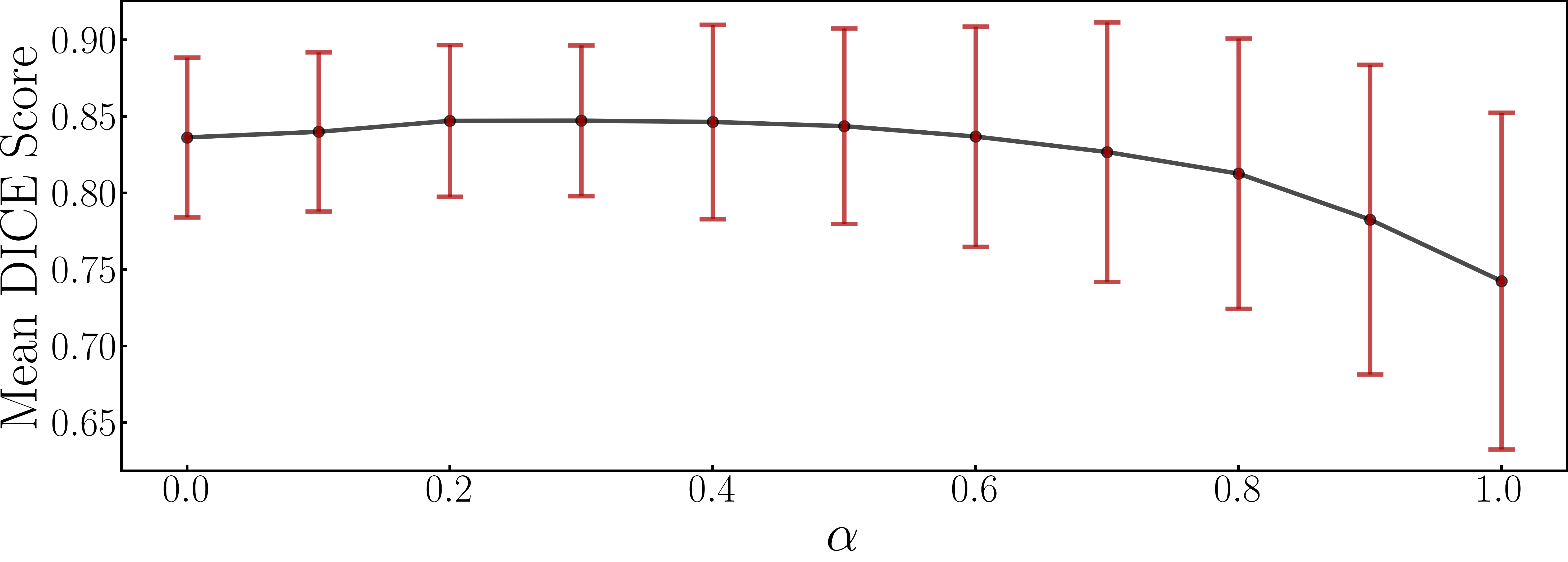}
    \caption{We evaluate how the weight of the feature-based distance loss term influences the registration results. 
    $\alpha = 0$ corresponds to the baseline and $\alpha = 1$ corresponds to \textit{variant 2}. Here, the DINOv2 encoder with the  cosine dissimilarity. The peak DSC is 0.847 $\pm$ 0.06 at $\alpha=0.3$ with 1.723 $\pm$ 1.018 HD95.}
    \label{fig:weighting}
\end{figure}

\subsubsection{Qualitative Comparison of General Vision Encoder Features}
We qualitatively compare the extracted features of one randomly chosen image between the three encoder models. Fig. \ref{fig:features-example} shows the major principal components of the features. 
The DINOv2-encoded features capture large structures with coarse outlines. 
SAM-encoded features appear to focus more on edges. 
Also, small structures are visible, and the features appear to be more fine-grained compared to the features of the DINOv2 encoder.
In contrast, the MedSAM-encoded features highlight the texture, but the edges are less apparent. 
Additionally, the DINOv2-encoded features of the background region are non-homogeneous, but this is not the case for the other two encoders. 

\subsubsection{Influence of the Feature Distance Weighting}
Furthermore, we evaluate how the parameter $\alpha$, which balances the image dissimilarity term $\mathcal{D_I}$ and the feature-based distance term $\mathcal{D_\mathcal{F}}$, influences the registration results for \textit{variant 2}. 
To this end, we vary $\alpha$ in steps of $0.1$ in the range of $[0.0, 1.0]$. 
To maintain the overall weighting between the distance measurement terms and the regularization term, we also weigh the image distance measure term $\mathcal{D}_I$ with $1-\alpha$, i.e., both weightings sum up to 1 in every case. 
For this experiment, we chose the DINOv2 encoder, as it showed the best combined result in terms of DICE and HD95 in Fig. \ref{fig:quantitative-results}. 
The results, with a peak DICE score of \(0.847 \pm 0.06\) with \(1.723 \pm 1.018\) HD95 at $\alpha=0.3$ is shown in Fig. \ref{fig:weighting}. 
We do not see a significant peak. However, it is clear that there is a decrease in performance once the weighting factor gets too extreme in any direction. 
This supports our finding that feature-based distances are not sufficient to drive the deformable registration task. 
An example registration with this best configuration is shown in Fig. \ref{fig:example-qualitative}. 
Here, we can see a plausible deformation in the region of the heart and an improved difference image.

\subsubsection{Influence of Image Upscaling}
In this experiment, we evaluate the effect of the upscaling factor on the final registration results. 
As described in Sec. \ref{sec:implementaiton-details}, we upsample the images before feeding them into the encoders. 
To achieve this, we perform registration with the DINOv2 encoder without prior upscaling of the images. 
This yields a mean DSC of \(0.683 \pm 0.122\) and a mean HD95 of \(3.664 \pm 1.658\). 
In comparison, with upscaling, a DSC of \(0.742 \pm 0.108\) and a HD95 of \(2.936 \pm 1.465\) is obtained. 
In both cases, the cosine dissimilarity and loss \textit{variant 1} were used. 
This indicates that upscaling the input images and, thus, obtaining feature maps with higher resolution results in an improved registration, as expected.

\section{Conclusion and Outlook}
In this work, we compared two variants of incorporating features from three general vision encoders in medical image registration. 
Our extensive evaluation on a cardiac cine MR dataset shows that using feature-based distance as an additional guidance term improves the registration.

A limitation of the general vision encoders is that they require the input images to be 2D images. For 3D volumes, a possible solution could be to stack multiple encoded 2D slices. However, this requires very large computing and memory resources. 
A disadvantage of our approach is the increased runtime with the repeated inference of the encoders. This could be circumvented by encoding the images once and performing the optimization in feature space. However, the meaning of a spatial deformation of high-dimensional feature maps is unclear. 

In future work, we plan to explore the use of feature maps from intermediate encoder levels and perform evaluation on a diverse range of datasets. Furthermore, we aim to extend our approach to 3D volumes and to compare them to 3D medical foundation models, such as UniGradICON \cite{tian2024unigradicon}.

\begin{credits}
\subsubsection{\ackname} This work was supported by the Federal Ministry of Education and Research (BMBF, Grant Nr. 01ZZ2315B and 01KX2021), the Bavarian Cancer Research Center (BZKF, Lighthouse AI and Bioinformatics), and the German Cancer Consortium (DKTK, Joint Imaging Platform).
\end{credits}

\bibliographystyle{splncs04}
\bibliography{bib}

\end{document}